# Multi-objective fluorescent molecule design with a data-physics dual-driven generative framework

Yanheng Li[1,2], Zhichen Pu[2], Lijiang Yang[1], Zehao Zhou[2,3†], Yi Qin Gao[1,4†]

1. New Cornerstone Science Laboratory, College of Chemistry and Molecular Engineering, Peking University, Beijing 100871, China.

2. Bytedance Seed, Beijing, China.

3. Zhongguancun Academy, Beijing, China.

4. Department of Chemistry, School of Science, Westlake University, 600 Dunyu Road, Hangzhou, Zhejiang Province 310030, P.R. China.

†To whom correspondence should be addressed.

E-mail: gaoyq@pku.edu.cn, zhouzehao@bjzgca.edu.cn

Author ORCIDs:
Yanheng Li         0000-0002-9643-9397
Zhichen Pu         0000-0003-3368-1052
Lijiang Yang       0000-0002-8175-0609
Zehao Zhou         0000-0003-2185-6855
Yi Qin Gao         0000-0002-4309-9376

# Abstract

Designing fluorescent small molecules with tailored optical and physicochemical properties requires navigating vast, underexplored chemical space while satisfying multiple objectives and constraints. Conventional generate-score-screen approaches become impractical under such realistic design specifications, owing to their low search efficiency, unreliable generalizability of machine-learning prediction, and the prohibitive cost of quantum chemical calculation. Here we present LUMOS, a data-and-physics driven framework for inverse design of fluorescent molecules. LUMOS couples generator and predictor within a shared latent representation, enabling direct specification-to-molecule design and efficient exploration. Moreover, LUMOS combines neural networks with a fast time-dependent density functional theory (TD-DFT) calculation workflow to build a suite of complementary predictors spanning different trade-offs in speed, accuracy, and generalizability, enabling reliable property prediction across diverse scenarios. Finally, LUMOS employs a property-guided diffusion model integrated with multi-objective evolutionary algorithms, enabling *de novo* design and molecular optimization under multiple objectives and constraints. Across comprehensive benchmarks, LUMOS consistently outperforms baseline models in terms of accuracy, generalizability and physical plausibility for fluorescence property prediction, and demonstrates superior performance in multi-objective scaffold- and fragment-level molecular optimization. Further validation using TD-DFT and molecular dynamics (MD) simulations demonstrates that LUMOS can generate valid fluorophores that meet various target specifications. Overall, these results establish LUMOS as a data-physics dual-driven framework for general fluorophore inverse design.

# Main

## Introduction

Fluorescent small molecules play a pivotal role in bioimaging[1,2], chemical sensing[3], and optoelectronics[4], distinguished by their structural tunability, environmental sensitivity, and biocompatibility[5,6]. However, the successful design of fluorescent molecules necessitates a delicate balance among multiple, sometimes conflicting objectives, such as tailored excitation/emission wavelengths, high photoluminescence quantum yields, and large Stokes shifts[7]. Traditional discovery paradigms, which typically rely on local chemical modifications of known scaffolds (e.g., rhodamines and BODIPYs) combined with extensive trial-and-error, are constrained by two inherent limitations: first, a heavy dependence on heuristic domain expertise and iterative, labor-intensive synthesis-characterization-screening cycles; and second, the restriction to fine-tuning existing architectures, which limits the exploration of broader chemical space and hinders the discovery of novel chemotypes with substantially improved properties[8].

In recent years, artificial intelligence (AI) has rapidly advanced molecular design, providing a promising route to overcome the limitations of traditional approaches[9,10]. For fluorescent molecule discovery, several AI-driven approaches have emerged: Sumita et al. combined a recurrent neural network (RNN) generator with Monte Carlo tree search (MCTS) to navigate candidate structures[11]; Han et al. employed a graph convolutional network (GCN) to predict stepwise molecular editing[12]; and Zhu et al. coupled reinforcement learning with a machine-learning property predictor for generative design and high-throughput screening[13]. Despite these advances, efficient inverse design of fluorophores remains challenging, largely due to two persistent bottlenecks, the first of which is ineffective exploration in discrete chemical space. Most existing approaches operate directly within a vast, discrete chemical space using a serial generate-score-screen workflow. This paradigm becomes increasingly impractical when multiple objectives, hard constraints, and scaffold-level novelty must be satisfied simultaneously[14]. The second bottleneck involves the insufficient alignment between data-driven prediction and physics-based evaluation. Neural network (NN) predictors often extrapolate unreliably to out-of-distribution (OOD) molecules[15]. In contrast, physics-based quantum chemical methods provide transferability and interpretability[16], yet their computational cost remains prohibitive for high-throughput use. Developing scalable strategies that integrate the merits of both is essential for discovering fluorophores with novel scaffolds, yet remains underexplored. Together, these limitations motivate a framework that can perform objective-conditioned generation while incorporating scalable, physics-anchored evaluation.

To address these challenges, we present LUMOS (Latent Unified fraMework for fluOrophore deSign), a data- and physics-driven framework for inverse design of fluorescent small molecules (Fig. 1). LUMOS couples generation and prediction modules through a shared latent representation, enabling efficient navigation of chemical space and direct specification-to-molecule mapping. To support diverse prediction scenarios, LUMOS combines NNs with an accelerated time-dependent density functional theory (TD-DFT) calculation workflow to yield a suite of complementary predictors, enabling accurate and efficient physics-anchored property assessment. Built on this representation, LUMOS enables controllable design through a latent diffusion model, supporting property-directed generation by prompt conditioning or by gradient-guided sampling using a differentiable predictor.

We systematically evaluated LUMOS across multiple dimensions, spanning representation learning, property prediction, and molecular design. Representation learning benchmarks show high reconstruction fidelity, accurately decoding latent representations back to their corresponding molecules while preserving key structural semantics. For fluorescence property prediction, LUMOS outperforms baseline models across multiple benchmarks, delivering high predictive accuracy together with physical plausibility and interpretability; moreover, integrating machine-learning predictors with physics-based TD-DFT calculations enables robust generalization to OOD molecules. Beyond prediction, LUMOS supports a broad range of design tasks from *de novo* generation to molecular optimization, enabling both scaffold-level exploration and local fragment refinement with consistent improvement over representative baselines. Validations using TD-DFT calculations and molecular dynamics (MD) simulations demonstrate that LUMOS can produce molecules with desirable target properties, highlighting the potential of LUMOS as a general framework for fluorescent molecule design.

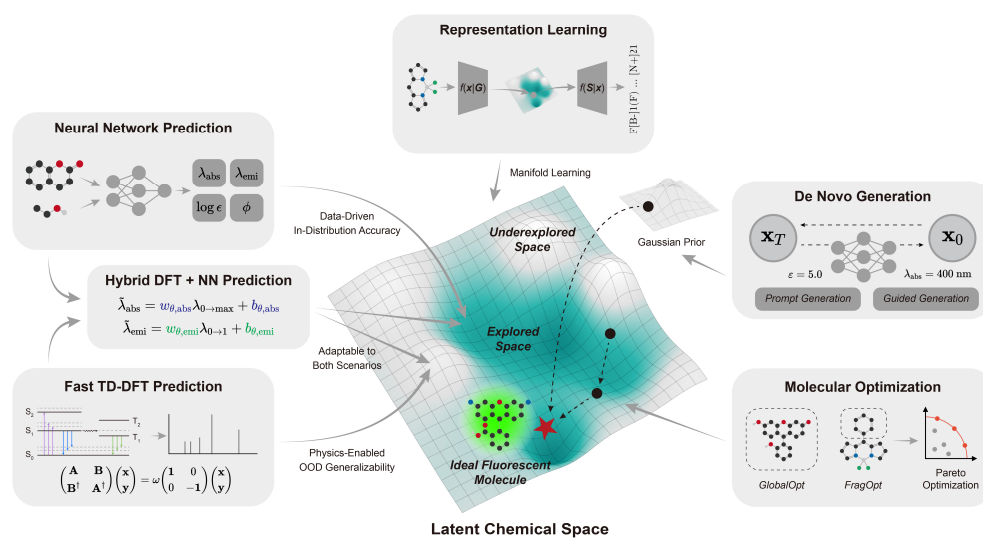

**Fig. 1 | Schematic overview of LUMOS.**
The workflow consists of three interconnected modules: representation learning, property prediction and molecule generation. **Top**: The representation learning module maps fluorescent molecules into a continuous latent chemical space (center), forming the navigational foundation of the entire workflow. **Left**: The prediction module leverages data-driven neural network (NN) models for rapid screening and navigating within explored regions, while a high-throughput TD-DFT pipeline ensures out-of-distribution (OOD) generalizability. A TD-DFT/NN hybrid predictor synergizes the strengths of both, unifying data-driven accuracy with physics-enabled generalizability. **Right**: The generation module exploits the latent space for two distinct tasks: *de novo* generation and molecular optimization, offering tailored modes for diverse design scenarios.

# Results

Construction of a continuous and semantic latent chemical space

To enable efficient navigation within the rugged landscape of fluorescence properties, we establish a mapping from the discrete, high-dimensional chemical space to a continuous and compact latent manifold. We adopted a graph-to-sequence autoencoder architecture from our previous work[17] and fine-tuned it on the FluoDB dataset[13] (details in Methods). As illustrated in Fig. 2a, molecules are first parsed with RDKit[18] into molecular graphs and encoded by a graph transformer model[19]. To unify molecules of varying sizes into a fixed-dimensional latent representation, we introduce virtual atoms as padding nodes; the compressed embeddings of these virtual nodes constitute the latent vectors. These latent vectors are then used as prefix tokens for a transformer decoder, which reconstructs the molecule in SMILES format. The model was trained via maximum likelihood estimation (MLE) with L2 regularization imposed on the latent vectors. This regularization constrains the topology of the latent manifold, preventing the model from merely memorizing and ensuring so that it captures meaningful, generalized structural semantics.

We further validated the learned representation across three progressive analyses. First, we evaluated the reconstruction accuracy, a basic requirement for a usable latent representation (Fig. 2b). The model achieved high reconstruction fidelity for both the in-distribution FluoDB test set (94.0%) and an external TADF dataset (77.8%), supporting generalization beyond the training distribution. Notably, when reconstruction fails, the model decoded inputs into valid molecules with closely related structures in most cases (> 80%), rather than producing invalid SMILES strings. This indicates that the learned manifold is continuous and compact, with few invalid regions. Second, we examined whether the latent vectors encode chemically meaningful semantics. Similarity analysis revealed a strong positive correlation between latent cosine similarity and chemical-space Tanimoto similarity (Fig. 2c), demonstrating that the manifold effectively preserves the structural similarity relationships. Finally, we compared our learned representations with those of the baseline model CDDD[20] (Continuous and Data-Driven Descriptors) via t-SNE analysis (Fig. 2d and Supplementary Fig. S1). Our model maps molecules with distinct fluorophore scaffolds into well-separated, compact clusters with clearer separation between fluorophore types, while the CDDD-derived representations exhibit significant scattering and inter-class entanglement. Collectively, these results confirm that the fine-tuned graph-to-sequence architecture effectively captures the intrinsic structural information of fluorescent molecules and provides a robust foundation for subsequent predictive and generative tasks. Details for training and validation are shown in Supplementary Section S1.

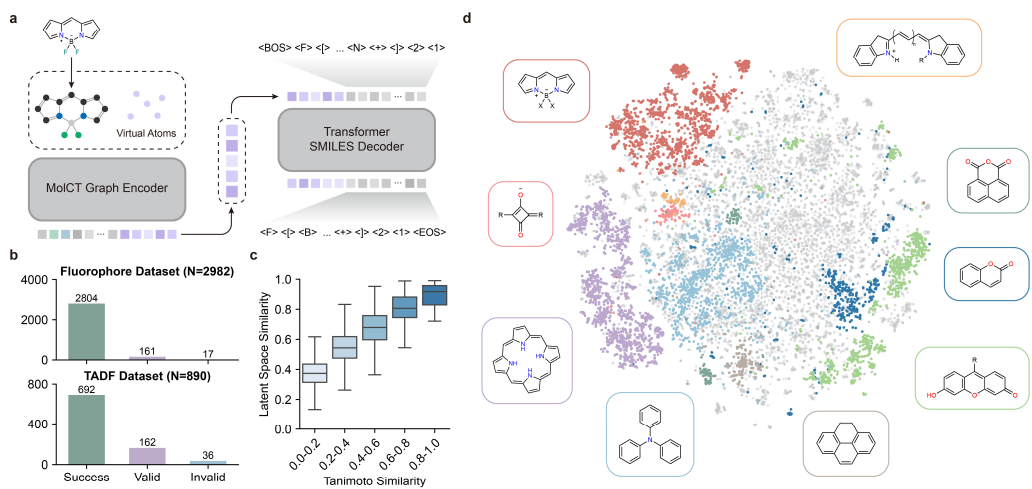

**Fig. 2 | The representation learning framework and analysis of latent chemical space.**
**a,** Model architecture of the graph-to-sequence autoencoder. Molecular graphs padded with virtual atoms are encoded by the MolCT Graph Encoder. The embeddings of virtual atoms are extracted to form a fixed-length latent representation, serving as prefix tokens to condition the transformer SMILES decoder for reconstruction.
**b**, Reconstruction accuracy evaluation on two datasets: an in-distribution fluorophore test set and an OOD TADF dataset. Bars indicate the counts of molecules that were perfectly reconstructed (Success), generated as valid but different SMILES (Valid), or generated as invalid strings (Invalid).
**c**, Correlation between chemical Tanimoto similarity and latent space cosine similarity based on 1,000 randomly sampled molecules in the FluoDB dataset, confirming that the latent space preserves chemical structure information. The boxes represent the interquartile range (IQR, 25th–75th percentile), the lines inside represent the median, and the whiskers represent data within ±1.5 × IQR.
**d**, Visualization of the fluorescent chemical space via t-SNE dimensionality reduction. The gray background represents the entire dataset, while colored clusters highlight distinct fluorophore scaffolds (shown in surrounding boxes).

## Accurate and physically consistent prediction of fluorescence properties

With the continuous and semantic latent chemical space constructed, we next sought to learn the mapping from the chemical space to the fluorescence properties, enabling efficient screening and providing guidance for generative exploration within the latent manifold. To this end, we developed a dual-branch prediction system (Fig. 3a) comprising an attentive graph predictor (AGP) for fast and interpretable property prediction and a latent surrogate predictor (LSP) that learns a direct latent-to-property mapping from the learned latent embeddings. To capture correlations between coupled properties and improve generalizability, both predictors employ a multi-task learning scheme to simultaneously regress four key properties: absorption maximum ($\lambda_{abs}$), emission maximum ($\lambda_{emi}$), (the logarithm of) molar extinction coefficient ($\log \epsilon$), and photoluminescence quantum yield (PLQY or $\phi$).

To evaluate the framework's predictive performance across diverse application scenarios, we assessed AGP and LSP under three data splitting strategies: random, scaffold, and a custom fluorophore split (Fig. 3b), where molecules are segregated based on their distinct fluorophore substructures (Supplementary Table S1). This split reduces fluorophore-level data leakage and provides a stringent test of the model's ability to generalize to unseen fluorescent chemotypes, better reflecting practical discovery settings. We benchmarked the regression root mean square error (RMSE) against two baseline methods: MPNN[21] and FLSF[13], a recently reported state-of-the-art (SOTA) model. As shown in Table. 1, AGP, MPNN and FLSF achieved comparable accuracy on random split for most properties, whereas AGP achieves the best overall performance on the scaffold and fluorophore splits. We attribute this improved extrapolation to the cross-attention design and multi-task training, which promote transferable structure-property relationships. LSP is slightly less accurate than AGP, likely because the autoencoder's reconstruction pre-training objective emphasizes global structural semantics and can underweight subtle functional-group variations that strongly affect fluorescence. Notably, under the same hidden dimensions ($d = 256$), the trainable parameters per predicted property for AGP (193,345) and LSP (342,913) are much fewer than MPNN (509,941) and FLSF (1,661,953), while achieving comparable or superior performance, supporting their efficiency for high-throughput screening.

Beyond benchmarking regression accuracy, we further examined the model's capacity to internalize physically meaningful structure–property relationships. We first visualized the atomic attention weights from the AGP for two representative molecular pairs where specific substitutions of electron donors or acceptors induce substantial shifts in fluorescence properties[7,22] (Fig. 3c). A striking alignment was observed between the learned attention weights and the density distributions of frontier molecular orbitals (HOMO/LUMO) calculated by DFT, demonstrating that AGP can autonomously identify key fluorophores and electron donating/withdrawing groups without explicit supervision on electronic structures. This ability to capture implicit electronic information is consistent with the stronger OOD performance of AGP. Next, we evaluated physical plausibility by examining the Stokes shift ($\Delta_{Stokes} = \lambda_{emi} - \lambda_{abs}$). Most fluorophores exhibit positive Stokes shifts (regular Stokes shifts) following Kasha's rule, while anti-Stokes shifts can occur under specific conditions (Fig. 3d). Accurately distinguishing between these modes is critical, as it reflects the model's grasp of the thermodynamic relationship between ground and excited states rather than treating absorption and emission as independent processes. To quantify this capability, we evaluated the four models across the three splits by the *Stokes error rate*, defined as the fraction of predictions in which the sign of the predicted shift mismatches the ground truth. As shown in Fig. 3e, both AGP and LSP achieved significantly lower error rates compared to baseline models. This physical robustness is attributed to our multi-task learning strategy, which enables the model to capture the joint structure-property dependence of $\lambda_{emi}$ and $\lambda_{abs}$.

Collectively, these analyses demonstrate that our framework achieves high predictive precision while maintaining physical plausibility. Furthermore, this dual-model architecture significantly extends the range of downstream applications. The lightweight and intrinsically interpretable AGP is well suited for rapid, high-throughput screening and mechanistic analysis, offering a substantial speed advantage over post-hoc attribution methods like SHAP analysis. Conversely, the differentiable LSP is essential for tasks that require gradient access, such as the gradient-guided diffusion strategy described in the subsequent section. Additional details for model architecture and evaluation are provided in Methods and Supplementary Section

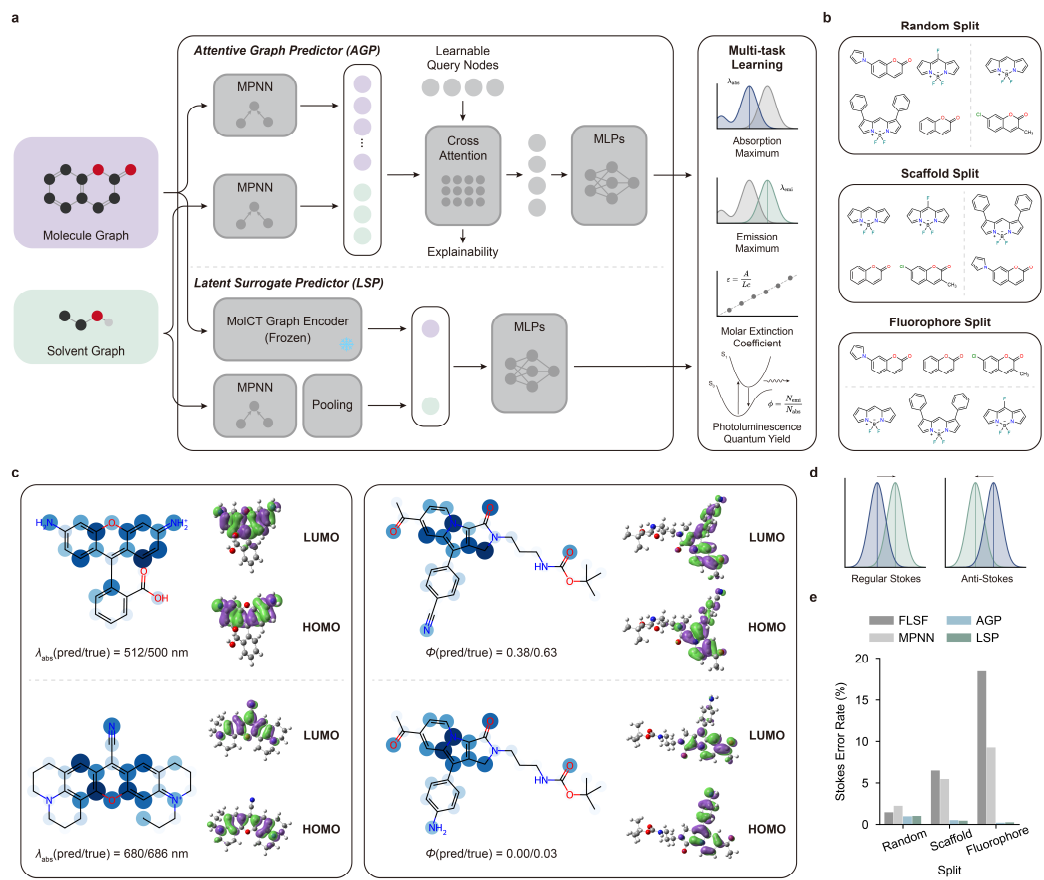

**Fig. 3 | Physically consistent and interpretable prediction of fluorescence properties.**
**a,** Architectures of the dual-branch prediction system and the multi-task learning scheme. Both predictors accept molecular and solvent graphs as input to simultaneously regress four photophysical properties. The attentive graph predictor (AGP, top) uses MPNN for molecule encoding and utilizes a cross-attention module to capture atomic contributions, while the latent surrogate predictor (LSP, bottom) uses the frozen MolCT encoder.
**b,** Illustration of the three data splitting strategies: random split, scaffold split (based on Murcko scaffolds), and fluorophore split (based on distinct fluorophore substructures).
**c,** Interpretability analysis comparing the AGP-learned attention weights with DFT-calculated molecular orbitals. Blue circles indicate the magnitude of atomic attention weights associated with the target properties annotated below ($\lambda_{abs}$ and $\phi$). The calculated HOMO/LUMO densities are shown on the right.
**d, e,** Evaluation of physical consistency using Stokes shift. **(d)** Schematic of regular Stokes shift (positive) and anti-Stokes shift (negative). **(e)** The Stokes error rate (the fraction of predictions where the sign of the predicted shift mismatches the ground truth) across different models.

**Table. 1 | Comparison of prediction RMSE of different models across three data splitting strategies.**

| Split | Model | Abs. | Emi. | LogE | PLQY |
|---|---|---|---|---|---|
| Random split | MPNN | 26.62 | 21.38 | 0.264 | 0.172 |
| | FLSF | 26.32 | 22.65 | 0.280 | 0.173 |
| | Attentive graph predictor | 26.80 | 23.30 | 0.266 | 0.186 |
| | Latent surrogate predictor | 27.56 | 25.36 | 0.290 | 0.188 |
| Scaffold split | MPNN | 57.13 | 48.56 | 0.449 | 0.266 |
| | FLSF | 63.98 | 57.14 | 0.471 | 0.285 |
| | Attentive graph predictor | 58.34 | 47.74 | 0.432 | 0.258 |
| | Latent surrogate predictor | 62.35 | 53.81 | 0.441 | 0.262 |
| Fluorophore split | MPNN | 48.62 | 49.45 | 0.327 | 0.342 |
| | FLSF | 57.91 | 56.58 | 0.334 | 0.356 |
| | Attentive graph predictor | 54.67 | 50.71 | 0.292 | 0.339 |
| | Latent surrogate predictor | 46.77 | 51.15 | 0.333 | 0.348 |

The table reports the root mean square error (RMSE) for four target properties: absorption maximum (Abs., nm), emission maximum (Emi., nm), logarithm of molar extinction coefficient (LogE, dimensionless), and photoluminescence quantum yield (PLQY, dimensionless). The performance is evaluated across three data splitting strategies: random split, scaffold split, and fluorophore split. MPNN and FLSF serve as baseline models.

Enhancing prediction generalizability via a physics-informed hybrid framework

Despite the robust performance of our graph-based predictors, purely data-driven approaches can be limited when extrapolating to underexplored chemical space. We observed that for OOD samples, the prediction RMSE of NN models can reach ~50 nm (Table. 1). This deviation spans roughly one-third of the visible spectrum, posing a substantial challenge for precise screening of high-performance fluorescent molecules. Quantum chemical calculations based on time-dependent density functional theory (TD-DFT) provide a physics-based alternative; however, its practical use for efficient prediction is constrained by two key bottlenecks. First, computational cost: standard workflows (e.g., Gaussian[23]) typically require hours to days per molecule, making them intractable for high-throughput evaluation. Second, systematic bias: raw TD-DFT calculations often exhibit significant deviations from experimental measurements, arising from approximations in exchange-correlation (XC) functionals and implicit solvation models[24].

To leverage quantum chemical calculations while achieving accuracy, generalizability, and practical throughput for fluorescence prediction, we developed a hybrid framework that combines a GPU-accelerated TD-DFT pipeline (Fig. 4a) with a bias prediction network (Fig. 4b). The TD-DFT calculation pipeline comprises three stages: first, initial conformations are generated and coarsely optimized with RDKit given the molecules and solvent dielectric constants ($\varepsilon$); second, the conformations are further optimized using the semi-empirical method xTB[25]; and finally, excitation spectra are computed using GPU4PySCF[26,27] (see Methods and Supplementary Section S3.1 for details). We estimate $\lambda_{abs}$ from the excitation wavelength from the ground state ($S_0$) to the maximum excitation state ($S_{max}$) and estimate $\log \epsilon$ using the corresponding oscillator strength. For emission, since our pipeline currently does not support excited-state geometry optimization, the vertical excitation wavelength from $S_0$ to the first excited state ($S_1$) is used to estimate $\lambda_{emi}$. This can be improved through the further development of fast excited-state optimization methods.

We benchmarked this workflow against standard Gaussian calculations using a test set of 49 molecules with varying sizes (see details in Supplementary Section S3.3). As shown in Extended Data Fig. 1 and Fig. 4c, our approach achieves accuracy comparable to Gaussian for $\lambda_{abs}$ and $\log \epsilon$. For $\log \epsilon$, both methods show limited agreement with experiment. This is because TD-DFT computes oscillator strengths, which are related to extinction coefficients through an integral over the absorption band rather than a direct equivalence. For $\lambda_{emi}$, our method aligns well with the Gaussian results calculated under the same protocol (without $S_1$ optimization), with noticeable deviations primarily in the long-wavelength regime; Gaussian calculations with $S_1$-optimized geometries yield the lowest RMSE. Notably, while maintaining accuracy comparable to Gaussian, our workflow accelerates calculations by approximately three orders of magnitude, reducing the per-molecule cost to ~$10^1$-$10^2$ seconds. This speedup is pivotal for scaling quantum chemical evaluations to large datasets.

However, as previously noted, systematic errors persist in TD-DFT: while the pipeline captures relative trends well (high $R^2$), it suffers from significant absolute deviations (high RMSE). To address this issue, we developed a bias prediction network (Fig. 4b). The network takes the molecular and solvent graphs as input and predicts the dynamic scaling ($w_\theta$) and shifting ($b_\theta$) factors, which are applied linearly to calibrate the raw TD-DFT outputs. This hybrid architecture harnesses the strengths of both paradigms: It preserves the extrapolation capability of physics-based calculations while leveraging the NN's capacity to correct complex biases.

We validated this approach on a subset of the fluorophore split test set ($n = 1,948$), benchmarking it against raw TD-DFT calculation and pure neural network (NN) (see details in Supplementary Section S3.3). As illustrated in Fig. 4d, the hybrid model (TD-DFT + NN) demonstrated significant improvements for $\lambda_{abs}$ and $\lambda_{emi}$ in both RMSE and $R^2$, consistently outperforming the standalone methods. Collectively, these results support the hybrid framework as a robust tool for high-throughput screening that offers improved accuracy and generalizability. In contrast, pure NN predictors (like AGP and LSP) retain a clear advantage in computational speed. Consequently, we propose a hierarchical screening strategy: pure NNs serve as just-in-time predictors for molecular optimization and as coarse filters for large datasets, while the hybrid framework acts as a subsequent tool for fine-grained post-hoc screening. This stratified approach forms the basis of our molecular optimization workflow introduced in the following section.

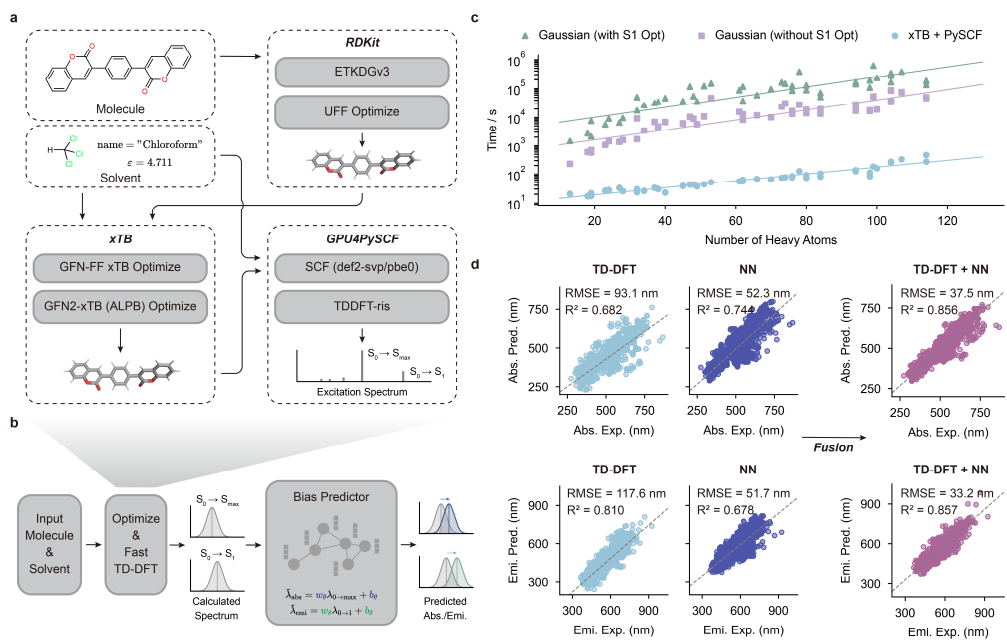

**Fig. 4 | Enhancing prediction accuracy and generalizability via a hybrid physics-informed framework.**

**a,** The high-throughput TD-DFT calculation workflow. The pipeline integrates RDKit for conformational search and initial optimization, xTB for multi-stage geometry optimization (considering implicit solvation), and GPU4PySCF for accelerated SCF and TD-DFT calculations to obtain the excitation spectrum.

**b,** Schematic of the hybrid TD-DFT/NN prediction model. A bias prediction network is used to correct the systematic errors of raw TD-DFT calculations. It takes molecular and solvent graphs as input to predict dynamic scaling ($w_\theta$) and shifting ($b_\theta$) factors, mapping the calculated excitation wavelengths ($S_0$ to $S_{max}$ for absorption, $S_0$ to $S_1$ for emission) to experimental measurements.

**c,** Computational cost comparison between standard Gaussian workflows and our accelerated pipeline. Our high-throughput workflow (blue circles) demonstrates orders of magnitude speedup compared to Gaussian (green triangles with excited-state optimization; purple squares without), showing favorable scaling with system size.

**d,** Parity plots comparing the prediction performance of raw TD-DFT, the pure neural network (NN), and the hybrid (TD-DFT + NN) predictor. Metrics including RMSE and $R^2$ are annotated in each subplot. The gray dashed lines represent the linear fitting results for each method.

## Targeted *de novo* generation of fluorescent molecules

The predictive framework establishes a forward mapping from chemical space ($\mathcal{M}$) to the property manifold ($\mathcal{P}$). However, fluorescent molecular design ultimately poses an inverse problem: deducing a molecule that satisfies specified property requirements. To address this, we developed a latent diffusion framework to model the conditional distribution $p(\mathcal{M}|\mathcal{P})$. Specifically, in the forward diffusion process, the molecular latent vector ($\mathbf{x}_0$) is progressively corrupted by Gaussian noise via a transition kernel $q(\mathbf{x}_t|\mathbf{x}_{t-1})$, converging to an isotropic Gaussian distribution ($\mathbf{x}_T \sim \mathcal{N}(\mathbf{0}, \mathbf{I})$) after $T$ steps. During the generation process, a diffusion transformer (DiT) model is trained to reverse this process by predicting the noise $\epsilon_\theta(\mathbf{x}_t, \mathbf{p}, t)$ conditioned on the target properties $\mathbf{p}$. By iteratively denoising via a backward kernel $p_\theta(\mathbf{x}_{t-1}|\mathbf{x}_t, \mathbf{p})$, the model reconstructs a valid latent representation, which is subsequently decoded into a SMILES string by the pre-trained decoder.

To address a broad spectrum of design scenarios, we implemented two conditioning mechanisms: prompt-conditioned generation and gradient-guided generation (Fig. 5a). In the prompt-conditioned mode, solvent information (dielectric constant, $\varepsilon$) and target properties ($\lambda_{abs}$, $\lambda_{emi}$, $\log \epsilon$ and $\phi$) are encoded via dedicated embedders. These conditions are injected into the DiT model through adaptive layer normalization (adaLN) module, a standard conditioning paradigm in generative models[28]. Conversely, gradient-guided generation couples an unconditional DiT with the differentiable LSP. Specifically, we define a loss function $\mathcal{L}$ based on the target fluorescence properties and use the differentiability of the LSP to compute the gradient $\nabla_{\mathbf{x}_t}\mathcal{L}$. Analogous to steered molecular dynamics[29], this gradient acts as an external biasing force that actively steers the denoising trajectory toward regions of the latent manifold associated with the desired properties (see details in Methods and Supplementary Section S4).

These two paradigms offer distinct advantages. Prompt-conditioned generation enables faster inference by eliminating the need for backpropagation during sampling. Furthermore, by parameterizing the solvent environment through the dielectric constant, it naturally extends to solvent mixtures and heterogeneous microenvironments when an effective dielectric constant is available. In contrast, gradient-guided generation prioritizes flexibility. Since the guidance relies on the external LSP, the system can be rapidly adapted to new data via retraining or fine-tuning. This modular design further enables customizable property targeting, as the objective function $\mathcal{L}$ can be defined to reflect arbitrary design objectives.

We rigorously evaluated both generative paradigms. For prompt-conditioned generation, we first assessed single-property control under two solvent environments: polar ($\varepsilon = 78.0$) and non-polar ($\varepsilon = 5.0$) (Fig. 5b). Target prompts were sampled from percentiles of the training distribution, and generated molecules were validated using TD-DFT and NN (Fig. 5b and Extended Data Fig. 2). We observed a strong positive correlation between input prompts and validated properties for $\lambda_{abs}$ and $\lambda_{emi}$ in both environments, highlighting the framework's potential for the precise engineering of desired optical signatures. For $\log \epsilon$ and PLQY, this correlation is weaker and often confined to a limited range. We primarily attribute this limitation to data scarcity, which constrains the model's capacity to learn the corresponding conditional distributions. Our total training dataset contains only ~45,000 molecule-solvent pairs with fluorescent labels; for individual properties the number of available pairs ranges from ~10,000 to ~20,000, and molecules with multiple concurrent labels are rarer still (only 5,041 molecules have all four labels). The impact of data sparsity is also reflected in the generative metrics (Supplementary Fig. S2): while chemical validity remains high, uniqueness and novelty decline, particularly when prompts are sampled from the distribution tails.

We further assessed multi-objective control by simultaneously prompting for $\lambda_{abs}$ and $\log \epsilon$ under non-polar conditions (Fig. 5c). TD-DFT validation indicates a strong correlation between the input prompts and the two-dimensional distribution of generated molecules, supporting the framework's capacity for multi-objective design.

Finally, we evaluated gradient-guided diffusion by generating molecules (in ethanol) with maximized values of four fluorescence properties. Compared to the unconditionally generated baseline (Fig. 5d), the molecules generated under guidance exhibit significantly enhanced property values. For $\log \epsilon$, while the NN predicted values indicate a substantial increase, TD-DFT validated oscillator strengths only show marginal differences. This discrepancy is likely attributable to the physical non-equivalence between oscillator strength and $\log \epsilon$. Collectively, these findings demonstrate the potential of the guided diffusion paradigm for the *de novo* generation of property-optimized molecules. Moreover, because the guidance loss can be defined flexibly, the guided mode readily extends to a broad range of generation objectives, as shown in the following section.

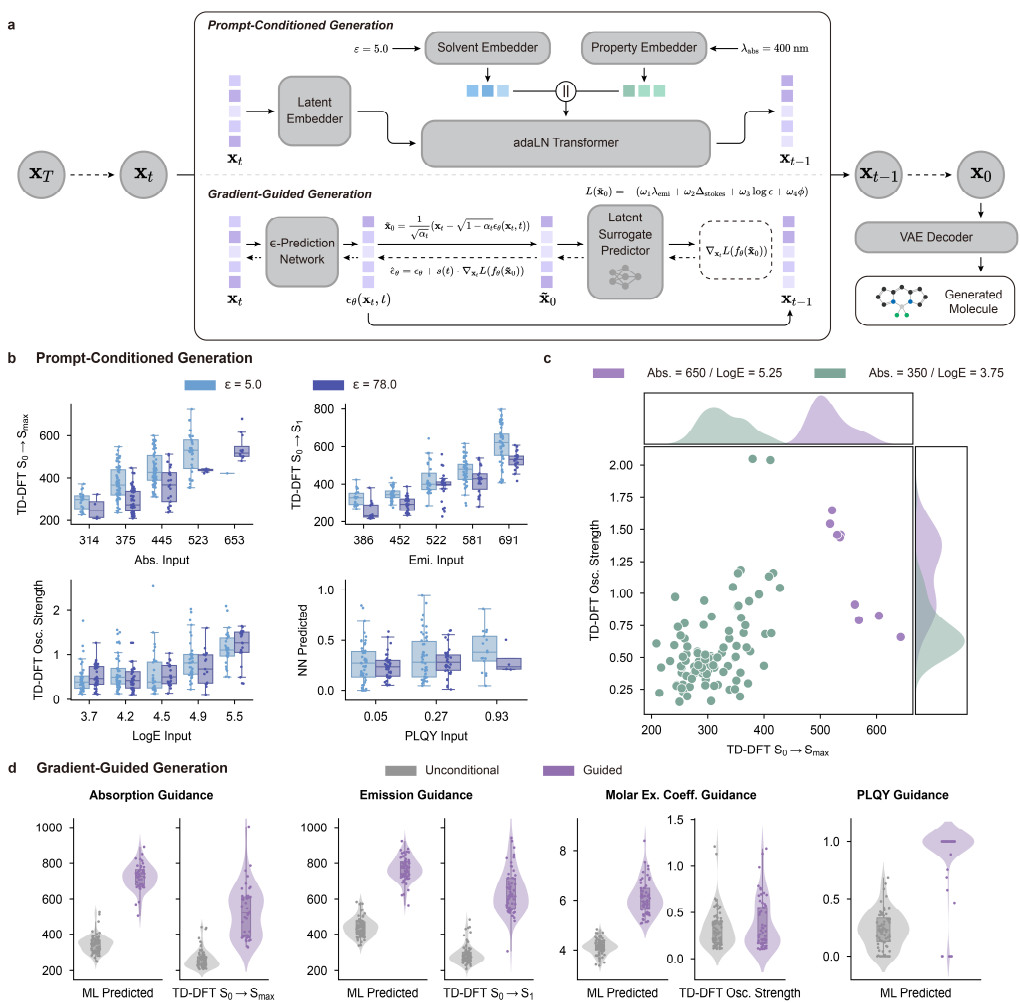

**Fig. 5 | Controllable *de novo* generation of fluorescent molecules via dual-mode diffusion.**
**a,** Schematic of the latent diffusion generative pipeline. The framework supports two control mechanisms: prompt-conditioned generation (top), which injects target properties and solvent dielectric constant $\varepsilon$ via adaptive layer normalization (adaLN); and gradient-guided generation (bottom), which utilizes gradients from the frozen latent surrogate predictor (LSP) to actively steer the denoising trajectory toward desired property landscapes.
**b, c,** Validation of prompt-conditioned generation. **(b)** Performance under single-property prompts. The boxplots demonstrate the correlation between input prompt values and validated properties for $\lambda_{abs}$, $\lambda_{emi}$, $\log \epsilon$, and PLQY under different solvent environments ($\varepsilon = 5.0$ and $78.0$). Validations are performed using TD-DFT (for $\lambda_{abs}$, $\lambda_{emi}$ and $\log \epsilon$) and NN (for PLQY). **(c)** Performance under dual-property prompts. The scatter plot shows the joint distribution of TD-DFT validated properties for molecules generated using two distinct dual-objective prompts, with a solvent environment of $\varepsilon = 5.0$; while the marginal kernel density estimation (KDE) plots on the top and right show the individual property distributions.

**d,** Assessment of gradient-guided generation. Violin plots compare the property distributions of molecules generated without guidance (unconditional, gray) versus those generated with gradient guidance (guided, violet) from the LSP. Experiments in this panel were simulated in an ethanol environment.

All boxes in this panel represent the IQR, the lines inside represent the median; whiskers extend to data within $1.5 \times$ IQR, and overlaid dots represent individual data points. For all TD-DFT validations, $\lambda_{abs}$ and $\lambda_{emi}$ correspond to the $S_0$ to $S_{max}$ and $S_0$ to $S_1$ excitation wavelengths, respectively, while the oscillator strength of $S_0$ to $S_{max}$ is used to estimate $\log \epsilon$.

## Versatile multi-objective molecular optimization for diverse design scenarios

Compared to *de novo* design, molecular optimization represents a more pragmatic strategy in fluorescent molecular engineering: starting from an initial molecule, molecules with improved properties are obtained through various modification strategies. This process typically necessitates balancing multiple objectives and constraints to obtain Pareto-optimal solutions in chemical space. To address this issue, we integrated the diffusion noise-denoising cycle into an evolutionary framework (Fig. 6a). Specifically, the noise-denoising procedure serves as a mutation operator that generates a population of structurally similar variants. Crucially, denoising is actively steered by predefined loss functions. The resulting offsprings are then selected using the NSGA-III algorithm[30]. After the evolution, candidate molecules undergo fine-grained screening using the hybrid model introduced above. This modular architecture is well suited for various optimization tasks by enabling flexible integration of property predictors, customizable selection criteria, and tailored guidance losses.

To evaluate the framework in realistic scenarios, we first applied it to a fluorescent probe optimization task. Ideal fluorescent probes typically require the following properties[7]: a long $\lambda_{emi}$ to enhance tissue penetration, a large Stokes shift to minimize self-quenching, and high brightness, reflected by large $\log \epsilon$ and PLQY. We started with a lead molecule (Fig. 6c) and optimized it for these four properties. Throughout the optimization process, we observed steady improvements across all metrics (Fig. 6b), with the values consistently surpassing those of the initial molecule (shaded regions). We then characterized representative optimized molecules (Fig. 6c). Validation via both LSP (NN) and the hybrid model (DFT + NN) confirmed consistent improvements in all objectives. Moreover, we benchmarked our framework against three baseline methods (Table 2): QMO[31] (a general multi-objective molecular optimization framework), Gen-DL[12] (a fluorescent molecule design framework), and REINVENT4[32] (a general molecular design framework, which has recently been used for fluorescent molecule design[13]). Our method outperformed these baselines, achieving higher hypervolume (HV) and improved target property values. It also showed advantages in optimization success rate and the number of valid molecules generated. Together, these results support the effectiveness of our framework for fluorescent molecular optimization.

In addition to global optimization, there is a frequent need for local fragment optimization, where the core scaffold remains fixed and specific fragments are modified. We further evaluated our framework on this task (Fig. 6d). Here, we partition each molecule into an optimizable fragment and a fixed core, and apply the noise-denoising process exclusively to the fragment. The generated fragments are reattached to the core at predefined anchors, and the assembled candidates are scored by the AGP and screened by NSGA-III to form the next generation. After evolution, the hybrid model is used for final screening. Analysis of the optimization trajectory (Fig. 6e) and representative final molecule (Fig. 6f) indicate consistent improvements across all four properties. Furthermore, comparison with the three baseline methods (Table 2) shows that our approach outperforms the baselines, demonstrating the framework's potential for local fragment optimization tasks.

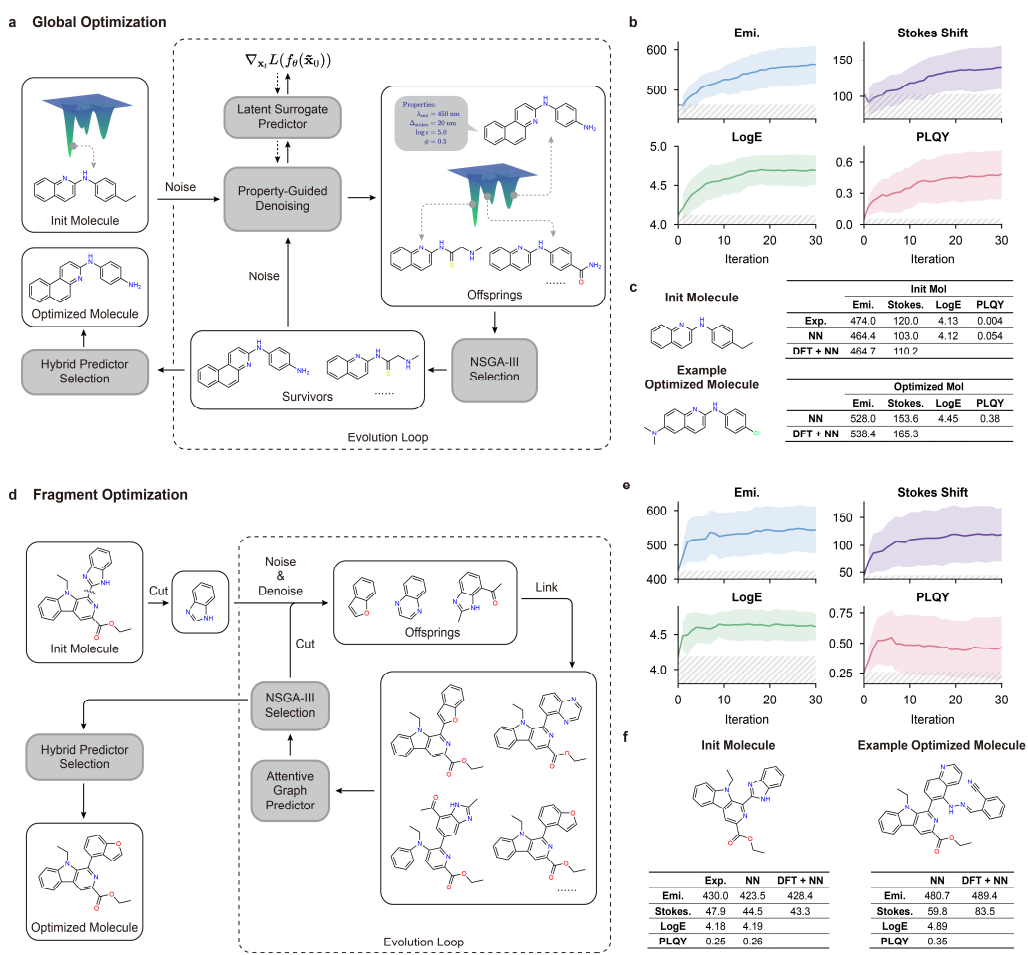

**Fig. 6 | LUMOS enables multi-objective molecular optimization at global and substructure scales.**
**a,** Schematic of the global optimization mode. This mode employs a noise-denoising mutation operator, utilizing gradients from the latent surrogate predictor (LSP) to steer the diffusion process for directed evolution. The NSGA-III algorithm is applied for selection to obtain a Pareto-optimal population.
**b,** Evolutionary trajectories of four target properties ($\lambda_{emi}$, Stokes shift, $\log \epsilon$ and PLQY) during global optimization. Solid lines represent the population mean at each iteration, while colored bands indicate the IQR. The hatched gray region denotes the property baseline of the initial molecule.

**c,** Structural and property comparison between the initial molecule and a representative global-optimized molecule. Properties validated by experiment (Exp.), predicted by both latent surrogate model (NN) and hybrid model (DFT + NN) are listed in tables.

**d,** Schematic of the fragment optimization mode. The workflow involves a predefined fixed core and a mutable fragment. The noise-denoising mutation is restricted to the fragment's latent representation, and the re-assembled offspring is evaluated by the attentive graph predictor (AGP) to guide NSGA-III selection.

**e,** Evolutionary trajectories during fragment optimization. The notation follows the same convention as in **(b)**.

**f,** Structural and property comparison between the initial molecule and a representative fragment-optimized molecule. Properties validated by experiment (Exp.), predicted by both attentive graph model (NN) and hybrid model (DFT + NN) are listed in tables.

**Table. 2 | Performance comparison of LUMOS and the baseline methods in molecular optimization tasks.**

| Task | Method | HV (↑) | Emi (↑) | Stokes (↑) | LogE (↑) | PLQY (↑) | #Mols (↑) | Success rate (NN) (↑) | Success rate (NN + DFT) (↑) |
|---|---|---|---|---|---|---|---|---|---|
| Global optimization | LUMOS | 0.814 | 670.57 | 225.20 | 5.23 | 0.89 | 793 | 52.2% | 8.1% |
| | QMO | 0.241 | 582.94 | 155.80 | 4.86 | 0.37 | 7 | 85.7% | 14.3% |
| | Gen-DL | 0.383 | 702.46 | 158.04 | 5.19 | 0.61 | 85 | 25.9% | 1.2% |
| | REINVENT4 | 0.315 | 610.18 | 178.31 | 4.86 | 0.49 | 2601 | 17.3% | 2.2% |
| Fragment optimization | LUMOS | 0.808 | 673.40 | 216.09 | 5.21 | 0.99 | 1241 | 61.3% | 50.3% |
| | QMO | 0.082 | 427.09 | 64.34 | 4.18 | 0.46 | 3 | 0.0% | 0.0% |
| | Gen-DL | 0.220 | 525.49 | 149.27 | 4.65 | 0.56 | 94 | 8.5% | 7.4% |
| | REINVENT4 | 0.518 | 582.94 | 194.33 | 4.94 | 0.87 | 13112 | 11.8% | 10.3% |

The table reports various comparative metrics for multi-objective fluorescent molecular optimization. HV: Hypervolume, a metric indicating the quality and diversity of the Pareto front. Emi., Stokes, LogE, and PLQY represent the population maximum values for emission maximum, Stokes shift, logarithm of molar extinction coefficient, and photoluminescence quantum yield, respectively. #Mols: Number of generated molecules that satisfy the constraints. Success rate: The percentage of generated molecules that Pareto-dominate the initial molecule in the objective space, evaluated under two models: pure neural network (NN, LSP for global optimization and AGP for fragment optimization) and the hybrid model (NN + DFT).

Finally, we address a common scenario in wet-lab fluorescent probe discovery. Sometimes a probe exhibits desirable fluorescence properties but fails in biological applications due to suboptimal ADMET characteristics, such as limited membrane permeability for intracellular imaging. The key challenge is to improve permeability-related properties without compromising the established fluorescent profile. Fluorescein exemplifies this dilemma: at physiological pH, it exists predominantly as an anion, which severely limits its membrane permeability[3].

Leveraging our framework's modularity, we integrated the ADMET-AI[33] predictor to optimize three permeability-related properties (Fig. 6a): PAMPA (parallel artificial membrane permeability assay), lipophilicity, and solubility. Simultaneously, we imposed constraints on three fluorescence properties: $\lambda_{emi}$, Stokes shift, and brightness (defined as $\log \epsilon \times \text{PLQY}$). This task presents a high-dimensional challenge: as the number of objectives and constraints increases, the search space grows exponentially. To address this, we customized the guided diffusion loss function to steer the denoising trajectory toward regions of chemical space that satisfy the predefined fluorescence thresholds, while retaining NSGA-III algorithm as the core engine for many-objective optimization.

The optimization results show consistent improvement in the three permeability-related properties (Fig. 7b). Crucially, fluorescence properties of the optimized molecules were maintained above the predefined thresholds and remained comparable to the input molecule (Fig. 7c). We further validated the starting molecule and three optimized candidates (Opt-1 to Opt-3) using quantum chemical calculations and molecular dynamics. First, we calculated the potential of mean force (PMF) free energy profiles for membrane translocation (Fig. 7d). As molecules move from the aqueous phase into the phospholipid bilayer, a lower PMF indicates a more thermodynamically favorable translocation process. The optimized molecules exhibit a stronger tendency for membrane translocation than Fluorescein. This trend is further supported by effective membrane permeability ($\log P_{eff}$) calculations[34] (Fig. 7e). While fluorescein shows a $\log P_{eff}$ of −8.90 (indicating impermeability), Opt-1 (−0.69), Opt-2 (−0.50), and Opt-3 (−1.87) show markedly improved permeability.

Notably, the model employed divergent optimization strategies to achieve this. Opt-1 and Opt-2 improve permeability by neutralizing ionizable groups (for example, via esterification or chlorination) to maintain neutrality at physiological pH. In contrast, Opt-3 carries a cationic charge and a long aliphatic chain; given its PMF rise near the membrane center (Fig. 7d), we suggest that it may function as a membrane-anchoring probe. These observations highlight the model's capability to explore diverse modification pathways and leverage multiple chemical mechanisms. Finally, evaluation via the hybrid model (Fig. 7g) reveals that Opt-1 to Opt-3 maintain or slightly improve upon the original fluorescence properties. Collectively, these results demonstrate the framework's flexibility and efficiency in tackling high-dimensional, complex optimization tasks in realistic design scenarios.

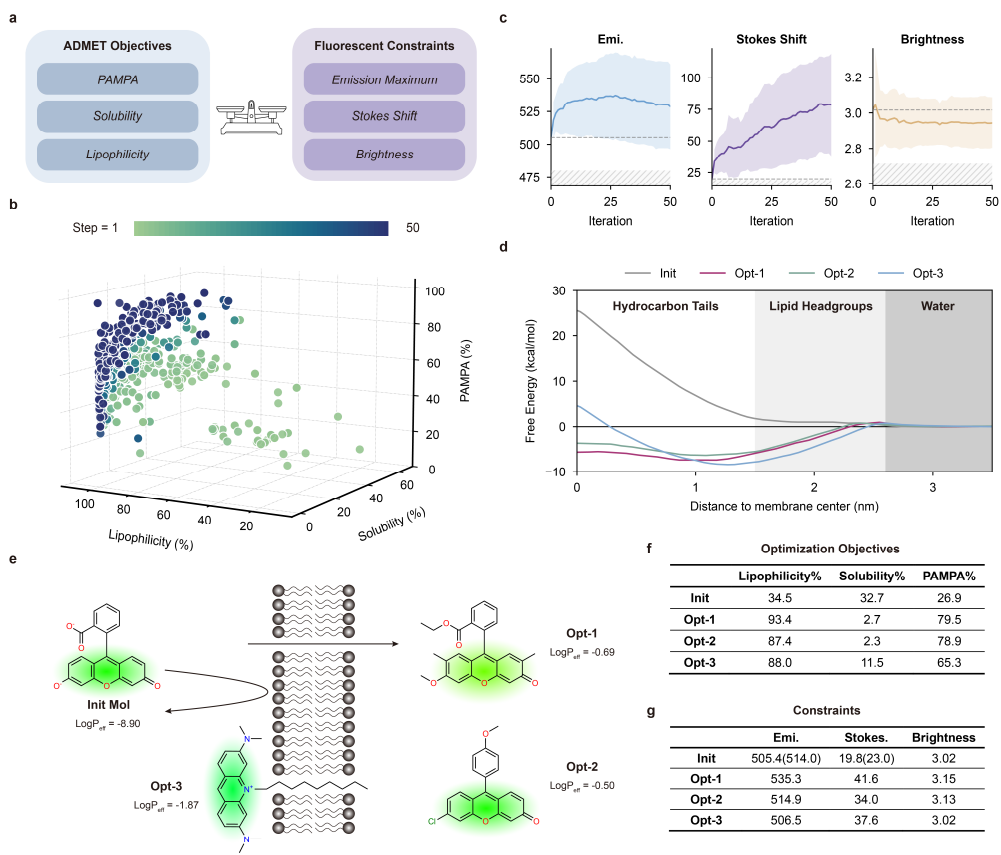

**Fig. 7 | Constrained multi-objective optimization for cell-permeable fluorophores.**

**a,** Schematic of the optimization setup balancing ADMET objectives against fluorescence constraints. The goal is to optimize ADMET properties (PAMPA permeability, solubility, lipophilicity) towards the Pareto frontier while maintaining fluorescence performance (emission, Stokes shift, brightness) above specific thresholds relative to the initial molecule.

**b,** 3D scatter plot visualizing the optimization trajectory in the ADMET objective space. Axes represent the predicted percentile scores by ADMET-AI. The color gradient (green to blue) indicates the evolutionary generation (Step 1 to 50).

**c,** Evolutionary trajectories of the constrained fluorescence properties. Solid lines represent the population mean, and colored bands indicate the IQR. The dashed line marks the property value of the initial molecule, while the hatched gray region denotes the violation zone (values below the constraint threshold).

**d,** Potential of mean force (PMF) free energy profiles for membrane translocation. Curves describe the free energy change as molecules move from bulk water (distance > 2.6 nm) to the center of a DOPC bilayer (distance < 1.5 nm), calculated via umbrella sampling and weighted histogram analysis method (WHAM).

**e,** Chemical structure and membrane interaction mechanisms of the initial molecule and three optimized candidates (Opt-1 to Opt-3), with the calculated cell permeability ($\log P_{\mathrm{eff}}$) listed below. Fluorophores of all molecules are colored with their emission wavelength. The initial molecule is not permeable, while optimized molecules can integrate into or cross the membrane.

**f, g,** Comparison of. **(f)** ADMET objectives and **(g)** Fluorescence constraints of the initial molecule and three optimized candidates. Experimental values for the initial molecule are provided in parentheses for validation.

## Discussion

Successful fluorescent molecule design requires searching vast, underexplored chemical space for one molecule that satisfies multiple objectives and constraints. In this regime, conventional generate-score-screen pipelines and data-driven methods often become insufficient or even impractical. In this work, we present LUMOS, a unified framework enabling efficient fluorophore inverse design through controllable, objective-guided generation and physically consistent prediction.

LUMOS is built on three conceptual pillars. First, a unified and navigable latent representation embeds discrete molecular graphs into a compact continuous manifold, providing a tractable space in which both sampling and optimization can be performed efficiently. Second, LUMOS leverages a diverse suite of predictors to accommodate distinct evaluation scenarios. The AGP adopts a lightweight architecture with a cross-attention design to give a fast and physically interpretable inference; the LSP enables backpropagation from target properties to molecular representations, which forms the basis of flexible objective-guided design. Complementing these data-driven components, the hybrid predictor combines quantum chemical excited-state calculations with machine-learning surrogates, achieving a practical balance among efficiency, generalizability, and accuracy. Third, a powerful generative engine coupled with multi-objective evolutionary algorithms supports the direct generation of candidates aligned with desired properties and enables iterative refinement to obtain Pareto-optimal molecules. This workflow naturally extends across practical cases, allowing both scaffold-level exploration and fine-grained refinement while balancing diverse objectives and constraints.

Notwithstanding these advances, several limitations warrant further investigation. First, prompt-conditioned generation currently exhibits limited novelty and uniqueness, and provides weaker control over certain targets (e.g., $\log \epsilon$ and PLQY), which is likely attributable to the scarcity of available datasets. Second, the current framework remains limited in modeling fluorophore behavior under complex environmental conditions, such as pH-responsive probes[35,36] or aggregation-induced emission (AIE) systems[37,38]. These effects are underrepresented in existing data resources, and general quantum chemical calculation protocols for such systems are still in early stages.

Future efforts should therefore prioritize improving fluorophore datasets, including expanding data volume and incorporating conformational information and environmental context. In parallel, deeper integration of quantum chemical approaches with AI-based methods will be important to extend inverse fluorophore design to increasingly realistic settings.

# Methods

Data preparation

**Autoencoder and diffusion transformer (DiT)**. The pre-training dataset for both the autoencoder and DiT consisted of a hybrid library of approximately 150 million molecules, aggregated and sanitized from ZINC[39], ChEMBL[40], and PubChem[41], following the procedures outlined in our previous work[17]. For fine-tuning, we used a curated dataset from FluoDB. Data sanitization involved the removal of molecules containing more than 128 heavy atoms, as well as those with multiple fragments, metal ions, or tautomers. For the autoencoder, the sanitized FluoDB dataset was randomly partitioned into training and test sets at a 9:1 ratio. Additionally, we utilized an external TADF test set[42], which underwent the same sanitization steps and was rigorously filtered to ensure no overlap with the training data, preventing any leakage. The training dataset for the DiT was constructed using the latent embeddings of molecules that were successfully reconstructed by the autoencoder.

**Fluorescent property prediction models**. Datasets for the AGP, LSP, and baseline models were derived from FluoDB. To comprehensively evaluate models, we partitioned the data using three distinct strategies: random split, scaffold split, and fluorophore split. For the random and scaffold splits, we adopted an 8:1:1 ratio for training, validation, and testing, respectively. For the fluorophore split, we selected BODIPY, coumarin, and naphthalimide derivatives as the test set, as these classes have well-defined scaffolds and sufficient sample sizes. The remaining molecules were randomly split at an 8:2 ratio for training and validation.

Representation learning framework

To encode fluorescent molecules into a compact, continuous latent space, we constructed an autoencoder architecture comprising two distinct components: a MolCT graph encoder $f_\theta(\mathbf{x}|\mathcal{G})$ and a transformer-based SMILES decoder $g_\phi(\mathcal{S}|\mathbf{x})$. We adopted this heterogeneous input-output architecture based on the premise that molecular graphs explicitly capture richer chemical information (e.g., hybridization and aromaticity), making them more suitable as encoder inputs; conversely, SMILES strings, organized as sequences, are intrinsically suited for autoregressive generation by the transformer decoder. The architecture, training, and inference details are described below.

**Graph Encoder.** The encoder employs a graph transformer architecture adapted from our previous work[19]. The input molecular graph $\mathcal{G}$ is first augmented with $P$ virtual atoms. Subsequently, an embedding module encodes the augmented graph into node feature vectors $\mathbf{H} \in \mathbb{R}^{(N+P) \times d}$ and edge feature vectors $\mathbf{E} \in \mathbb{R}^{(N+P) \times (N+P) \times d}$, where $N$ denotes the number of real atoms and $d$ represents the hidden layer dimension. The node vectors are updated via an attention module followed by a transition module:

$$\mathbf{Q}, \mathbf{K}, \mathbf{V}, \mathbf{B} = f_{\text{emb}}^{(l)}(\mathbf{H}^{(l)}, \mathbf{E}^{(l)}) \tag{1}$$

$$\mathbf{H}_{\text{int}}^{(l)} = \phi_{\text{out}}^{(l)}\left(\text{Softmax}\left(\frac{\mathbf{Q}\mathbf{K}^{\text{T}} + \mathbf{B}}{\sqrt{d}}\right)\mathbf{V}\right) \tag{2}$$

$$\mathbf{H}^{(l+1)} = f_{\text{transition}}\left(\mathbf{H}_{\text{int}}^{(l)}\right) = \phi_c^{(l)}\left(\sigma\left(\phi_a^{(l)}\left(\mathbf{H}_{\text{int}}^{(l)}\right)\right) \cdot \phi_b^{(l)}\left(\mathbf{H}_{\text{int}}^{(l)}\right)\right) \tag{3}$$

where $l$ denotes the layer index. $f_{\text{emb}}$ represents the attention embedding module (a collection of dense layers) used to project $\mathbf{H}^{(l)}$ and $\mathbf{E}^{(l)}$ into query, key, and value matrices $\mathbf{Q}, \mathbf{K}, \mathbf{V} \in \mathbb{R}^{(N+P) \times d}$ and the attention bias $\mathbf{B} \in \mathbb{R}^{(N+P) \times (N+P)}$. $\phi_{\text{out}}^{(n)}$, $\phi_a^{(n)}$, $\phi_b^{(n)}$, $\phi_c^{(n)}$ denote a series of dense modules, and $\sigma$ is the activation function. Note that the attention mechanism shown above is simplified for clarity; the actual implementation utilizes multi-head attention.

Following the node update, the edge vectors interact with the updated node vectors via an outerproduct module and a transition module. The outerproduct update is defined as:

$$\mathbf{C}_i, \mathbf{C}_j = \phi_{\text{in}}^{(l)}(\mathbf{H}^{(l+1)}) \tag{4}$$

$$\mathbf{E}_{\text{int}}^{(l)} = \phi_{\text{out}}^{(l)}\left(\text{Flatten}(\mathbf{C}_i \otimes \mathbf{C}_j)\right) \tag{5}$$

$$\mathbf{E}_{\text{int}}^{(l+1)} = f_{\text{transition}}\left(\mathbf{E}_{\text{int}}^{(l)}\right) \tag{6}$$

where $\phi_{\text{in}}^{(n)}$ and $\phi_{\text{out}}^{(n)}$ are projection dense modules, $\mathbf{C}_i, \mathbf{C}_j \in \mathbb{R}^{(N+P) \times c}$ are intermediate vectors, and Flatten denotes vectorization operation of a matrix. Note that for clarity, the update formulas presented above omit the residual connections. In practice, all encoder layers follow the ResiDual transformer design[43], incorporating residual connections between layers via a set of accumulate vectors and utilizing layernorm to ensure numerical stability as network depth increases. After $N_{\text{enc}}$ layers, we extract the node vectors corresponding to the virtual atoms ($\mathbf{P} \in \mathbb{R}^{P \times d}$). These are projected via a dense module $\phi_{\text{proj}}$ and normalized using L2 normalization to yield the final latent space representation $\mathbf{X} \in \mathbb{R}^{P \times h}$:

$$\mathbf{X} = \text{L}^2\text{Norm}\left(\phi_{\text{proj}}(\mathbf{P})\right) \tag{7}$$

**SMILES Decoder.** The decoder follows a standard ResiDual transformer architecture. During training, the input consists of the latent vector $\mathbf{X}$ and the one-hot encoding of the corresponding SMILES string. These are projected to the same dimension via an embedding module and concatenated into a unified hidden vector. After $N_{\text{dec}}$ layers of updates, the output logits $\mathbf{L} \in \mathbb{R}^{M \times V}$ are decoded via Softmax to obtain the probability $\mathbf{S}$ of the SMILES tokens. The model is optimized using cross-entropy loss:

$$\mathbf{S} = \text{Softmax}(\mathbf{L}) \tag{8}$$

$$\mathcal{L} = -\sum_i \mathbf{T}_i^{\text{T}} \log \mathbf{S}_i \tag{9}$$

Where $M$ is the length of the SMILES string, $V$ is the vocabulary size, $i$ is the token index, and $\mathbf{T}_i \in \mathbb{R}^V$ represents the one-hot encoding of the $i$-th SMILES token. During inference, the decoder receives the latent vector $\mathbf{X}$ and generates the molecule autoregressively. To prevent the decoding process from being trapped in local likelihood optima, we use beam search for sampling.

Prediction framework

In this work, we established a comprehensive framework to predict fluorescence properties, encompassing both neural networks (NNs) and a hybrid physics-informed model. The NN branch includes the attentive graph predictor (AGP) and the latent surrogate predictor (LSP). Both models take molecular and solvent graphs as input but serve distinct purposes: the AGP offers rapid inference and interpretable electronic distribution analysis, while the LSP provides a direct mapping from the latent space to properties, enabling differentiability for gradient-guided generation. Complementarily, the hybrid TD-DFT/NN model synergizes quantum chemical calculation with data-driven corrections, achieving superior OOD accuracy for $\lambda_{abs}$ and $\lambda_{emi}$, thus facilitating high-fidelity screening. The network architectures and the TD-DFT workflow are described below.

**Attentive graph predictor (AGP).** In AGP, both molecule and solvent graphs are encoded by MPNN, resulting in atom-level features $\mathbf{H}_{mol} \in \mathbb{R}^{N_m \times d}$ and $\mathbf{H}_{sol} \in \mathbb{R}^{N_s \times d}$. To capture the atomwise contribution to fluorescence properties, we introduced a cross-attention module with learnable query vectors. Specifically, for the $i$-th target property, we initialize a learnable query vector $\mathbf{q}_i \in \mathbb{R}^d$, where $d$ is the hidden dimension. The query vector $\mathbf{q}_i$ interacts with the key and value vectors $\mathbf{K}, \mathbf{V} \in \mathbb{R}^{(N_m + N_s) \times d}$ derived from the concatenation of molecule and solvent features, aggregating global information into a property-specific representation $\mathbf{f}_i \in \mathbb{R}^d$, which is then mapped to the property value $p_i$ via an MLP head $\phi_{pred}$:

$$\mathbf{K}, \mathbf{V} = f_{emb}(\mathbf{H}_{mol} \parallel \mathbf{H}_{sol}) \tag{10}$$

$$\mathbf{f}_i = \mathrm{Softmax}\left(\frac{\mathbf{q}_i \mathbf{K}^T}{\sqrt{d}}\right) \mathbf{V} = \mathbf{w}_i^T \mathbf{V} \tag{11}$$

$$p_i = \phi_{pred}(\mathbf{f}_i) \tag{12}$$

where $\parallel$ denotes the concatenation operation, and $f_{emb}$ is the attention embedding module (consistent with the graph encoder's definition in previous section). During inference, the attention weights $\mathbf{w}_i$ provide interpretability by highlighting atomic contributions. The mean squared error (MSE) loss is used as the objective function for the AGP (as well as for the LSP and hybrid model described below).

**Latent surrogate predictor (LSP).** In the LSP, the solvent graph is encoded using MPNN, while the molecular graph is processed by the frozen pre-trained MolCT graph encoder. Since the latent representation $\mathbf{X}$ has a fixed length, we apply a mean pooling operation to the solvent features $\mathbf{H}_{sol}$ to match the dimensionality. The two vectors are then concatenated and fed into an MLP for property regression:

$$p_i = \phi_{pred}\big(\mathrm{Flatten}(\mathbf{X}) \parallel \mathrm{Mean}(\mathbf{H}_{sol})\big) \tag{13}$$

This architecture ensures that the predictor is strictly aligned with the generative latent manifold, enabling effective gradient guidance.

**High-throughput TD-DFT workflow.** In this work, we developed a high-fidelity prediction framework combining high-throughput TD-DFT calculations with NN calibration. The TD-DFT calculation pipeline consists of three stages: (1) Conformer generation: Using RDKit, hydrogen atoms are added to molecules, then conformers are generated via the ETKDGv3 algorithm and pre-optimized by UFF force field (max iterations = 500). (2) Geometry optimization: Conformers undergo two-stage optimization using xTB, first via the traditional GFN-FF force field[44], followed by the semi-empirical GFN2-xTB[45] method with ALPB[46] implicit solvation model. (3) Excitation spectrum calculation: SCF and TD-DFT calculations are performed by GPU4PySCF. Since the systems we calculated were all closed-shell singlet molecules, considering the trade-off between computational accuracy and efficiency[24], we employed the PBE0 functional and def2-svp basis set, coupled with the IEF-PCM implicit solvation model. TDDFT-ris[47–49] method is used to compute the first 5 excited states (including excitation energies and oscillator strengths). To rigorously benchmark accuracy and efficiency, comparative calculations including ground state geometry optimization, vertical excitation, and excited-state ($S_1$) optimization were also conducted using Gaussian 16, with same the functional and basis set.

**Bias correction network.** From the excitation spectra, we extract the maximum absorption wavelength $\lambda_{0\to\max}$ and the vertical emission wavelength $\lambda_{0\to 1}$. These raw values are calibrated by an MPNN-based bias predictor. The network encodes the molecular and solvent graphs (with average pooling) and uses MLPs to predict two pairs of correction parameters: a scaling factor $w_\theta$ and a shifting factor $b_\theta$. The final calibrated predictions $\tilde{\lambda}_{\text{abs}}$ and $\tilde{\lambda}_{\text{emi}}$ are computed as:

$$\tilde{\lambda}_{\text{abs}} = w_{\theta,\text{abs}} \cdot \lambda_{0\to\max} + b_{\theta,\text{abs}} \tag{14}$$

$$\tilde{\lambda}_{\text{emi}} = w_{\theta,\text{emi}} \cdot \lambda_{0\to 1} + b_{\theta,\text{emi}} \tag{15}$$

Generative framework

To enable *de novo* conditional generation and multi-objective molecular optimization, we constructed a diffusion model on the latent manifold and integrated it into an evolutionary algorithm framework. Details of the latent diffusion model, prompt-conditioned generation, gradient-guided generation and optimization frameworks are described as below.

**Latent diffusion model**. We formulated the molecular generation task as a conditional denoising process within the continuous latent manifold learned by the autoencoder. The framework consists of a forward diffusion process and a backward generative process. For forward process, given a latent vector $\mathbf{x}_0$ derived from the pre-trained encoder, the forward process is a Markov chain that progressively injects Gaussian noise according to a variance schedule $\beta_1, \cdots, \beta_T$:

$$q(\mathbf{x}_t|\mathbf{x}_{t-1}) = \mathcal{N}(\mathbf{x}_t; \sqrt{1-\beta_t}\mathbf{x}_{t-1}, \beta_t \mathbf{I}) \tag{16}$$

Using the reparameterization trick, $\mathbf{x}_t$ at any timestep $t$ can be sampled directly from $\mathbf{x}_0$:

$$\mathbf{x}_t = \sqrt{\bar{\alpha}_t}\mathbf{x}_0 + \sqrt{1-\bar{\alpha}_t}\boldsymbol{\epsilon}, \qquad \boldsymbol{\epsilon} \sim \mathcal{N}(\mathbf{0}, \mathbf{I}) \tag{17}$$

where $\alpha_t = 1 - \beta_t$ and $\bar{\alpha}_t = \prod_{s=1}^{t} \alpha_s$. As $t \to T$, the distribution of $\mathbf{x}_T$ approaches a standard isotropic Gaussian $\mathcal{N}(\mathbf{0}, \mathbf{I})$. For the backward process, we used a parameterized kernel $p_\theta(\mathbf{x}_{t-1}|\mathbf{x}_t, \mathbf{p})$ to iteratively reconstruct the initial latent vector $\mathbf{x}_0$ from the noise $\mathbf{x}_T$. This transition is defined as a conditional Gaussian distribution:

$$p_\theta(\mathbf{x}_{t-1}|\mathbf{x}_t, \mathbf{p}) = \mathcal{N}(\mathbf{x}_{t-1}; \boldsymbol{\mu}_\theta(\mathbf{x}_t, t, \mathbf{p}), \sigma_t^2 \mathbf{I}) \tag{18}$$

where $\mathbf{p}$ represents the property condition. We parameterize the mean $\boldsymbol{\mu}_\theta$ by training a

neural network $\epsilon_\theta$ to predict the noise component:

$$\boldsymbol{\mu}_\theta(\mathbf{x}_t, t, \mathbf{p}) = \frac{1}{\sqrt{\alpha_t}}\left(1 - \frac{\beta_t}{\sqrt{1-\bar{\alpha}_t}}\epsilon_\theta(\mathbf{x}_t, t, \mathbf{p})\right) \quad (19)$$

The training objective is the Frobenius norm between the predicted and actual noise:

$$\mathcal{L}_{\text{diff}} = \mathbb{E}_{\mathbf{x}_0, t, \epsilon}\|\epsilon - \epsilon_\theta(\mathbf{x}_t, t, \mathbf{p})\|^2 \quad (20)$$

**Diffusion transformer (DiT) architecture and prompt-conditioned generation.** In this work, a diffusion transformer (DiT) model is used to predict the noise. The model accepts the noisy latent $\mathbf{x}_t$, timestep $t$ and property conditions $\mathbf{p}$ (including $\lambda_{\text{abs}}$, $\lambda_{\text{emi}}$, $\log \epsilon$, PLQY and solvent dielectric constant $\varepsilon$) as inputs. Scalar conditions are first min-max normalized (see details in Supplementary Section S4.1) and then vectorized via a set of $K$ Gaussian radial basis functions (RBFs):

$$e_k = \exp(-\gamma(p - \mu_k)^2) = \exp\left(-\frac{1}{K}\left(p - \frac{k}{K}\right)^2\right) \quad (21)$$

where $p$ is the normalized property, $e_k$ is the $k$-th element of the embedding $e \in \mathbb{R}^K$, and $\gamma, \mu_k$ are hyperparameters set to $\gamma = 1/K$ and $\mu_k = k/K$. To inject these conditions, we utilized adaptive layer normalization (adaLN). In this mechanism, $t$ and $\mathbf{p}$ are projected by MLPs to regress the scaling $\eta$ and shifting $\zeta$ parameters of the normalization layers, dynamically modulating the features ($\mathbf{h}$):

$$\text{adaLN}(\mathbf{h}, \mathbf{p}, t) = \eta(\mathbf{p}, t) \odot \text{LayerNorm}(\mathbf{h}) + \zeta(\mathbf{p}, t) \quad (22)$$

**Gradient-guided generation.** Besides prompt-conditioned generation, we also implemented a guided denoising framework for flexible property control[50]. In this mode, a tailored loss function $\mathcal{L}$ is constructed based on design objectives and the predicted noise $\epsilon_\theta$ is steered via the following formulas:

$$\hat{\mathbf{x}}_0 = \frac{\mathbf{x}_t - \sqrt{1-\bar{\alpha}_t}\epsilon_\theta(\mathbf{x}_t, \mathbf{p}, t)}{\sqrt{\bar{\alpha}_t}} \quad (23)$$

$$\hat{\epsilon}_\theta = \epsilon_\theta + s(t) \cdot \nabla_{\mathbf{x}_t}\mathcal{L}(\hat{\mathbf{x}}_0) \quad (24)$$

$$\delta = \arg\min_\delta \mathcal{L}(\hat{\mathbf{x}}_0 + \delta) \quad (25)$$

$$\hat{\epsilon}_\theta = \hat{\epsilon}_\theta - \sqrt{\frac{\bar{\alpha}_t}{1-\bar{\alpha}_t}}\delta \quad (26)$$

where $s(t)$ is a time-dependent scaling schedule controlling the guidance strength. To ensure sample diversity and stability, a resampling strategy is introduced, involving multiple noise-denoising loops at each step.

For *de novo* generation benchmarks (Fig. 4d), the loss function is defined as $\mathcal{L}_i = (m_i - p_i)^{q_i}$, where $p_i$ is the predicted property, and $m_i, q_i$ are hyperparameters designed to prevent invalid generation that deviates the reliable prediction region of LSP. Details for hyperparameter settings are shown in Supplementary Section S4.2.

**Molecular optimization.** By integrating latent diffusion with evolutionary algorithms, we address the multi-objective molecular optimization across different scales and scenarios.

Specifically, we employ a partial noise-denoising strategy acting as the mutation operator. The input molecule is corrupted for $T_{\text{opt}}$ steps (where $T_{\text{opt}} < T$) and then denoised in parallel to generate molecular population that retains structural similarity to the parent. The population is then selected via the NSGA-III algorithm to obtain Pareto-optimal offspring. As shown in our previous work[17], the choice of $T_{\text{opt}}$ depends on input molecules and task requirements. Details for $T_{\text{opt}}$ and other parameter settings for molecular optimization tasks are shown in Supplementary Section S5.1.

For global optimization task, we utilized the gradient-guided denoising method to guide the generated trajectories towards a molecular space with desirable properties. The loss function is defined as:

$$\mathcal{L}_{\text{global}} = -u(\lambda_{\text{emi}} + \lambda_{\text{abs}} + \log \epsilon + \phi) \tag{27}$$

where $u$ is an envelope function to penalize OOD samples (see details in Supplementary Section S5.1).

For fragment optimization, the molecule is split into a fixed core and a mutable fragment. Partial noise-denoising is applied to the fragment's latent vector. Since the LSP requires holistic molecular representation and cannot provide gradients for isolated fragment latents, we use the AGP for evaluation. Offspring fragments are decoded and assembled with the core by single bonds. Specifically, we defined the connectable sites in fragments as carbon or nitrogen atoms within the conjugated bonds that have at least one implicit hydrogen. The assembled molecules are scored by AGP to guide the NSGA-III selection process. It's worth noting that, for different application scenarios, we can use various connection rules, which also reflects the flexibility of our framework.

For cell permeability optimization, we also used the guided denoising approach to ensure the generated molecules satisfy the predefined constraints, where the loss function is defined as below:

$$\mathcal{L}_{\text{admet}} = -(\min(\lambda_{\text{emi}}, \rho_{\text{emi}}) + \min(\Delta_{\text{Stokes}}, \rho_{\text{Stokes}}) + \min(b, \rho_{\text{b}})) \tag{28}$$

where $\rho$ represents the constraint thresholds (using the related property values of the starting molecule), $b = \phi \cdot \log \epsilon$ is brightness. Different with global optimization, the gradient guidance vanishes if the constraints are already satisfied. Moreover, the permeability-related objectives (lipophilicity, solubility, PAMPA) are predicted using ADMET-AI.

Molecular dynamics calculation

To quantitatively evaluate cell permeability, we calculated the potential of mean force (PMF) and effective permeability ($\log P_{\text{eff}}$). The computational pipeline consists of system modeling, molecular dynamics (MD) simulation, free energy calculation, and permeability calculation, as described below.

**System modeling**. All simulation systems were constructed using the CHARMM-GUI[51] web server. The membrane model consisted of a DOPC bilayer with 72 lipids (36 per leaflet) aligned perpendicular to the z-axis, hydrated by a water layer of 22.5 Å thickness. For fluorescent molecules, the protonation states corresponding to pH = 7.4 were determined first and placed to the center of the membrane. CHARMM36m[52] force field was adopted for lipids,

CGenFF[53] for small molecules, and the TIP3P[54] model for water. The system charges were neutralized with Na$^+$ and Cl$^-$ ions, resulting in a total system size of approximately 20,000 atoms.

**MD simulation**. All simulations were performed using GROMACS[55] in the NPT ensemble with a time step of 2 fs. Bond lengths involving hydrogen atoms were constrained using the LINCS algorithm. The temperature was maintained at 310 K using the Berendsen thermostat (with $\tau_T = 1.0$ ps). The pressure was controlled at 1 bar using the Parrinello-Rahman barostat (with $\tau_p = 1.0$ ps and compressibility $4.5 \times 10^{-5}$ bar$^{-1}$). Long-range electrostatic interactions were treated using the Particle Mesh Ewald (PME) method, while van der Waals and Coulomb cutoffs were set to 1.2 nm. Following energy minimization and pre-equilibration, we generated initial configurations for umbrella sampling by pulling the fluorescent molecule from the membrane center to the bulk water along the z-axis using steered molecular dynamics (SMD).

**Free energy and permeability calculation.** We used umbrella sampling to enhance conformational sampling across the membrane. The distance between the center of mass (COM) of the molecule and the DOPC bilayer along the z-axis was chosen as the collective variable (CV). Sampling windows were spaced at 0.1 nm intervals, each restrained by a harmonic potential with a force constant of 1000 kJ mol$^{-1}$·nm$^{-2}$. Each window was simulated for a minimum of 60 ns. To ensure convergence for relaxation time in high-viscosity regions (specifically the lipid headgroup interface), simulations were adaptively extended up to 260 ns. The first 20 ns of each trajectory were discarded as equilibration. PMF profiles were reconstructed using the weighted histogram analysis method (WHAM) implemented in GROMACS and symmetrized assuming bilayer symmetry. Settings of simulation parameters were adapted from Bennion et al.[34]

The effective membrane permeability ($\log P_{\text{eff}}$) of a small molecule can be calculated by combining information from the PMF profile with diffusion coefficients in the membrane. Specifically, the diffusion coefficients $D(z)$ are calculated via the hummer's method[56]:

$$D(z) = \frac{Var(z)}{\tau_z} \quad (29)$$

where $Var(z)$ is the variance of the solute's COM within the umbrella sampling window, and $\tau_z$ is the characteristic time of the autocovariance decay. $\tau_z$ was determined by fitting the position autocorrelation function $\langle z(t)z(0)\rangle$ to a single exponential decay function $C(t) = \exp(-t/\tau)$. Having obtained the diffusion coefficient $D(z)$ and the PMF profile $\Delta G(z)$, the local resistance to permeation $R(z)$ and the total effective permeability $P_{\text{eff}}$ are calculated as[57]:

$$R(z) = \frac{\exp(\Delta G(z)/k_B T)}{D(z)} \quad (30)$$

$$\frac{1}{P_{\text{eff}}} = R_{\text{eff}} = 2 \int_0^Z R(z)dz \quad (31)$$

where $k_B$ is the Boltzmann constant, $T$ is the temperature (310 K), and $Z$ is the distance from the bilayer center to the bulk water phase.

# Data Availability

## Code Availability

The source code of LUMOS is available at https://github.com/egg5154/LUMOS.

## Acknowledgements

We sincerely acknowledge financial support from the National Natural Science Foundation of China (T2495221) and the New Cornerstone Science Foundation (NCI202305). Y.L. acknowledges Hao Li and Tai Wang for helpful discussion.

## Author Contributions Statement

Y.Q.G. and Z.Z. developed overall concepts in this study. Y.Q.G., Z.Z., Z.P. and L.Y. supervised the project. Y.L. designed the LUMOS framework and wrote the initial draft of this manuscript. All authors contributed ideas to the work and assisted in manuscript editing and revision.

## Competing Interests Statement

The authors declare no competing interest.

# Extended Data Figs

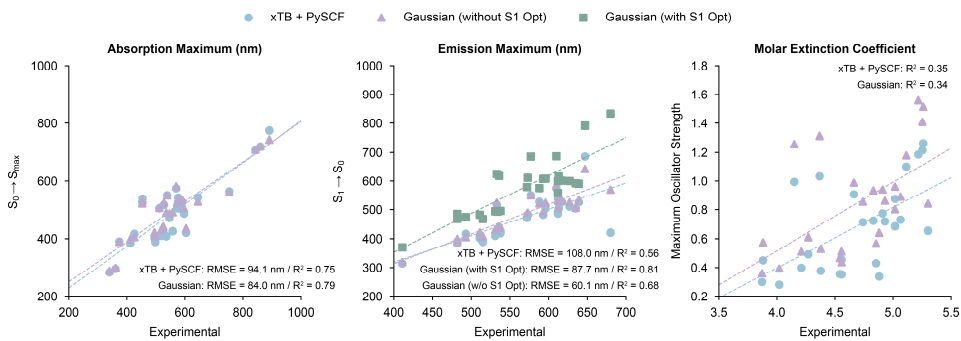

**Extended Data Fig. 1 | Accuracy comparison between the high-throughput workflow and standard Gaussian calculations.**
Scatter plots correlate experimental measurements with TD-DFT calculations for three key properties: absorption maximum ($\lambda_{abs}$, left), emission maximum ($\lambda_{emi}$, center), and molar extinction coefficient ($\log \epsilon$, right). PLQY is excluded due to the lack of a standardized TD-DFT calculation protocol. Dashed lines indicate linear regression results. Performance metrics (RMSE and $R^2$) are annotated in each subplot to quantify the alignment between calculated and experimental values. The calculated values for $\lambda_{abs}$, $\lambda_{emi}$ and $\log \epsilon$ correspond to the $S_0 \rightarrow S_{max}$ transition wavelength, $S_0 \rightarrow S_1$ transition wavelength and maximum oscillator strength, respectively. For $\lambda_{emi}$, results with $S_1$ optimization were also calculated using Gaussian.

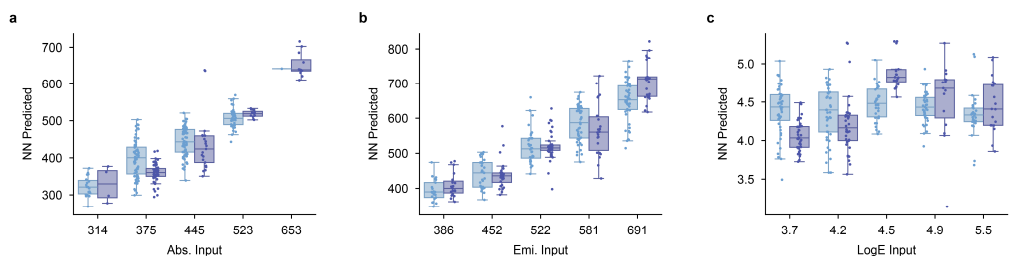

**Extended Data Fig. 2 | Validation of prompt-conditioned generation by neural network predictor.**
The boxplots illustrate the correlation between input target prompts and the properties of generated molecules as predicted by the AGP for **(a)** $\lambda_{abs}$, **(b)** $\lambda_{emi}$ and **(c)** $\log \epsilon$ under different solvent environments: $\varepsilon = 5.0$ (light blue) and 78.0 (dark blue). Boxplots indicate the median (center line), IQR (box), and data within 1.5 × IQR (whiskers), with overlaid dots representing individual data points.

# Supplementary Materials

# Contents

# Supplementary Information

## S1 Autoencoder Model

**S1.1 Model Hyperparameters and Training Protocol**

The model architecture are detailed in Methods. For the MolCT encoder, we utilized atom and bond feature dimensions of 256 and 128, respectively. The interaction module employed 3 interaction units with 8 attention heads. The continuous latent space was defined by 16 virtual nodes, with each node having a latent embedding dimension of 32. For the SMILES decoder, the model was configured with a feature dimension of 384, with 8 layers and 12 attention heads.

The training process consisted of two stages: pre-training and fine-tuning. In the pre-training stage, the autoencoder was trained using a mixed dataset comprising ZINC[1], ChEMBL[2], and PubChem[3], with a batch size of 256 for a total of 500,000 steps. In the fine-tuning stage, the model was trained using both the pre-training dataset and the FluoDB dataset[4], with the batch size set to 128 for each dataset and a total of 500,000 steps. The dropout rate was set to 0.01. The learning rate was initialized at 2e-4 and decayed to 2e-5 following a cosine schedule. The AdamW optimizer was used with a weight decay factor of 1e-4. Both pre-training and fine-tuning stages were performed on 8 NVIDIA A100 GPUs.

**S1.3 Validation Experiments**

The validation experiments comprise three components: reconstruction accuracy test, t-SNE dimensionality reduction analysis, and a comparison between molecular Tanimoto similarity and latent vector cosine similarity.

**Reconstruction accuracy test.** We utilized the test set from FluoDB (n = 2,982) and an external TADF dataset (n = 890). All molecular SMILES underwent the same sanitization protocol as the training set before being processed by the autoencoder. Decoding was performed using beam search with a beam size of 4. The reconstruction results were categorized as follows: 'Success' if the decoded SMILES exactly matched the input; 'Valid' if the decoded SMILES represented a chemically valid molecule distinct from the input; and 'Invalid' if the output did not correspond to a valid molecular structure.

**t-SNE dimensionality reduction.** We employed the entire FluoDB dataset for this analysis. Dimensionality reduction was performed using t-SNE initialized with PCA. In the visualization, fluorophore scaffolds were colored based on substructure matching using RDKit[5]; the corresponding SMARTS patterns are listed in Table S1. We compared our method with a baseline model, CDDD[6], using the identical dimensionality reduction protocol.

**Similarity comparison.** We randomly sampled 1,000 molecules from the dataset used in the t-SNE analysis and calculated their pairwise similarity matrices (extracting the lower triangular elements). Tanimoto similarity was calculated using Morgan fingerprints with a radius of 2 and a bit vector length of 2048.

## S2 Neural Network Predictors

**S2.1 Model Hyperparameters and Training Protocol**

For the attentive graph predictor (AGP), the model employs two message passing neural networks

(MPNNs) to independently encode the fluorescent molecule and solvent graphs. Both MPNNs are configured with a hidden dimension of 256 and a depth of 10 layers, with parameter sharing applied across layers. For the latent surrogate predictor (LSP), the fluorophore encoder utilizes the frozen MolCT encoder from the pre-trained autoencoder, resulting in a molecular embedding dimension of 512. The solvent is encoded using a separate MPNN with a hidden dimension of 256.

All predictors were trained using the FluoDB dataset. Detailed procedures for data cleaning and splitting are provided in the Methods section. The models were trained with a batch size of 64 for a total of 100,000 steps. The dropout rate was set to 0.1. The learning rate was initialized at 1e-3 and decayed to 1e-4 following a cosine schedule. The AdamW optimizer was employed with a weight decay factor of 0.01. Training was performed on a single NVIDIA A100 GPU.

**S2.2 Benchmarking**

We benchmarked the performance of AGP and LSP against two baseline methods: (1) MPNN, a JAX implementation of the message passing neural network architecture used in Chemprop[7]; and (2) FLSF[4], a fluorescence property predictor that integrates MPNN with fluorophore fingerprints. To ensure a fair comparison, the baseline models were configured with the same architectural hyperparameters (depth = 10, hidden dimension = 256) and were trained independently for each of the four fluorescence properties ($\lambda_{abs}$, $\lambda_{emi}$, $\log \epsilon$, and PLQY). The training schedule and optimizer settings were identical to those used for AGP and LSP.

To quantify the physical plausibility of the predictions, we compared the Stokes error rate across the four models. The Stokes error rate is defined as the fraction of predictions where the sign of the predicted Stokes shift mismatches the ground truth (i.e., predicting a negative 'anti-Stokes' shift when the experimental shift is positive, or vice versa):

$$\text{Stokes Error Rate} = \frac{1}{N} \sum_{i=1}^{N} \mathbb{I}(\text{sign}(\widehat{\Delta}_{\text{Stokes}}^{(i)}) \neq \Delta_{\text{Stokes}}^{(i)}))$$

For this evaluation, the test dataset was constructed from the intersection of the absorption ($\lambda_{abs}$) and emission ($\lambda_{emi}$) test sets used in the accuracy benchmarks.

**S2.3 Physical Interpretability Tests**

To evaluate the physical interpretability of the AGP, we compared the learned attention weights with the frontier molecular orbitals (FMOs, especially the HOMO and LUMO) derived from DFT calculations. Test molecules were selected from two works[8,9]. Initial 3D conformers were generated using RDKit. We then performed a two-stage geometry optimization using the xTB package[10]:

1. **Preliminary Optimization**: Conducted using the GFN-FF[11] force field with the command:

xtb --gfnff --opt normal

2. **Refined Optimization**: Performed using the GFN2-xTB[12] method with the ALPB[13] implicit solvation model:

xtb --gfn2 --alpb {solvent} --opt tight

Following optimization, orbital calculations were performed using the Gaussian[14] software package. We employed the PBE0 functional (PBE1PBE) with the def2-SVP basis set. Solvent effects were included via the SCRF method. The specific keywords used were:

#p pbe1pbe/def2svp pop=Regular gfinput Geom=Check Guess=Read scrf(solvent={solvent})

Solvent environments were selected to be consistent with the original literature: acetonitrile was used for the molecule pair in the absorption ($\lambda_{abs}$) prediction, and methanol was used for the pair in the PLQY prediction. Molecular orbitals were visualized using GaussView[15].

# S3 Hybrid Predictor

**S3.1 Calculation Details**

Details regarding the high-throughput TD-DFT calculation workflow are provided in the Methods section. The specific commands and parameters used for xTB optimization were identical to those detailed in Section S2.3. For Gaussian calculations, we performed three sequential steps for each molecule:

1. **Ground-State Geometry Optimization**:

#p pbe1pbe/def2svp opt freq scrf(solvent={solvent})

2. **TD-DFT Vertical Excitation**:

#p pbe1pbe/def2svp TD(NStates=5) opt Geom=Check Guess=Read scrf(solvent={solvent})

3. **TD-DFT Excited-State Optimization**:

#p pbe1pbe/def2svp TD(NStates=5) opt Geom=Check Guess=Read scrf(solvent={solvent})

All TD-DFT calculations are performed starting from the optimized ground-state conformation from Step (1). In rare instances, TD-DFT calculations failed to converge, resulting in either termination errors or predicted spectra that deviated drastically from experimental values. Such molecules were excluded from the subsequent statistical analyses. In Gaussian, we used different computational configurations for molecules of different sizes:

| #Atoms | #CPUs | Memory Allocation (GB) |
|---|---|---|
| < 30 | 8 | 16 |
| 30 ~ 150 | 16 | 32 |
| 150 ~ 300 | 32 | 64 |

**S3.2 The Bias Prediction Network**

The Bias Prediction Network employs two MPNNs to independently encode the molecular and solvent graphs. The resulting node representations are aggregated via mean pooling and subsequently processed by Multilayer Perceptrons (MLPs) to predict the systematic bias, which is defined as the residual between the TD-DFT calculated values and the experimental ground truth.

For the molecular graph encoder, the hidden dimension was set to 256. For the solvent graph encoder, the hidden dimension was set to 32. Both MPNN encoders were constructed with a depth of 10 layers. The training hyperparameters and optimization schedule were identical to those used for the AGP and LSP models, as detailed in Section S2.1.

**S3.3 Evaluation Datasets**

We first compared the computing speed of our high-throughput workflow against standard Gaussian calculations. Given the high computational cost associated with Gaussian, we constructed a representative subset by randomly sampling based on heavy atom counts from the fluorophore split test set. This yielded a sample of 50 molecules, of which 49 were successfully calculated. Next, we benchmarked the accuracy of the hybrid model against both pure TD-DFT calculations and pure neural network predictions (via AGP). Due to the computational cost, we filtered the fluorophore split test set to select molecules possessing complete experimental labels for $\lambda_{abs}$, $\lambda_{emi}$ and $\log \epsilon$. This resulted in a dataset of 1,973 molecules. Within this subset, 1,948 molecules successfully converged after the

TD-DFT calculation workflow.

# S4 Generation Framework

**S4.1 Model Hyperparameters and Training Protocol**

The latent diffusion transformer was constructed with an intermediate hidden dimension of 512. The network architecture consisted of a stack of 12 transformer layers, with each layer employing 16 attention heads to facilitate the multi-head attention mechanism within the latent space.

For conditional generation, the conditioning signals were first normalized using min-max scaling and subsequently expanded via 256 Gaussian Radial Basis Functions (RBFs). The normalization formula is defined as:

$$\varepsilon_{\text{normed}} = \frac{\varepsilon}{100}, \quad \lambda_{\text{normed}} = \frac{\lambda - 200}{900}, \quad \log \epsilon_{\text{normed}} = \frac{\log \epsilon - 0.5}{8.0}$$

Where $\varepsilon$ is dielectric constant, $\lambda$ is the absorption/emission wavelength, and $\log \epsilon$ is the (logarithm of) molar extinction coefficient.

The model was trained on a refined dataset, excluding molecules from both the pre-training and FluoDB datasets that failed autoencoder reconstruction. Training was performed with a batch size of 128 for a total of 1,000,000 steps on 8 NVIDIA A100 GPUs. A dropout rate of 0.01 was applied to the model weights. For molecules in the FluoDB dataset, property labels and solvent dielectric constants were incorporated as conditioning signals. During training, these conditions were randomly masked (dropped) with a probability of 0.1. The optimizer configuration and learning rate schedule were identical to those employed for the autoencoder (as detailed in Section S1.1).

**S4.2 Evaluation Details**

For each experimental setting, we generated 128 molecules. From this set, only unique and novel molecules were selected for further analysis. To validate the generation molecules, the properties for generated molecules were calculated via TD-DFT and the AGP model.

For gradient-guided generation, the generative process was steered by minimizing a specific loss function, defined as:

$$\mathcal{L}_i = (m_i - p_i)^{q_i}$$

where $p_i$ is the predicted property, and $m_i, q_i$ are hyperparameters designed to prevent invalid generation that deviates the reliable prediction region of LSP. The specific values for these hyperparameters are listed in the following table:

| Property | $m_i$ | $q_i$ |
|---|---|---|
| $\lambda_{\text{abs}}$ | 900 | |
| $\lambda_{\text{emi}}$ | 1000 | 2 |
| $\log \epsilon$ | 7 | |
| PLQY | 1 | 1 |

# S5 Molecular Optimization

**S5.1 Settings for Optimization Tasks**

The detailed settings for the three optimization tasks, including global optimization, fragment optimization, and cell permeability optimization, are summarized in the following table. Key parameters include the number of optimization steps ($N_{opt}$), population size ($N_{pops}$), number of offspring generated per molecule ($N_{offs}$), starting molecules, constraints (and their thresholds), noising schedule, and the solvent environment.

| Setting | Global Optimization | Fragment Optimization | Cell Permeability Optimization |
|---|---|---|---|
| $N_{opt}$ | 30 | 30 | 50 |
| $N_{pops}$ | 128 | 128 | 128 |
| $N_{offs}$ | 8 | 8 | 8 |
| Starting Molecule | CCc1ccc(Nc2ccc3ccccc3n2)cc1 | c1ccc2[nH]cnc2c1 | O=C(O)c1ccccc1-c1c2ccc(=O)cc-2oc2cc(O)ccc12 |
| Noising Schedule | Uniform[140, 160] | Uniform[140, 170] | Uniform[200, 220] |
| Solvent | CCO | ClC(Cl)Cl | O |
| Optimization Objectives | $\lambda_{emi}$, $\Delta_{Stokes}$, $\log \epsilon$ and PLQY | $\lambda_{emi}$, $\Delta_{Stokes}$, $\log \epsilon$ and PLQY | Lipophilicity, Solubility and PAMPA_NCATS |
| Constraints (Thresholds) | Tanimoto Similarity ($> 0.4$) | Synthetic Accessibility ($< 3.0$) | Has Substructure (SMARTS: [#6]1~[#6]~[#6]~[#6]2~[#6]~[#6]3:[#6]:[#6]:[#6]:[#6]:[#6]:3~[*]~[#6]~2~[#6]~1) $\lambda_{emi}$ ($> 0.95 \times \rho_{emi}$) $\Delta_{Stokes}$ ($> 0.95 \times \rho_{Stokes}$) Brightness = $\log \epsilon \times$ PLQY ($> 0.90 \times \rho_{brightness}$) |

Here, $\rho$ represents the property value of the starting molecule. For the global optimization task, we employed gradient-guided diffusion. The loss function is defined as:

$$\mathcal{L}_{global} = -u(\lambda_{emi} + \lambda_{abs} + \log \epsilon + \phi)$$

Note that the property values have been normalized using min-max scaling. The term $u$ serves as an envelope function designed to penalize out-of-distribution (OOD) samples, defined as follows:

$$u(x) = \begin{cases} -10(x-2) + 2 & x > 2 \\ x & -0.25 < x < 2 \\ 10(x + 0.25) - 0.25 & x < -0.25 \end{cases}$$

In practice, the envelope function was applied exclusively to $\lambda_{abs}$ and $\lambda_{emi}$.

# Supplementary Figures

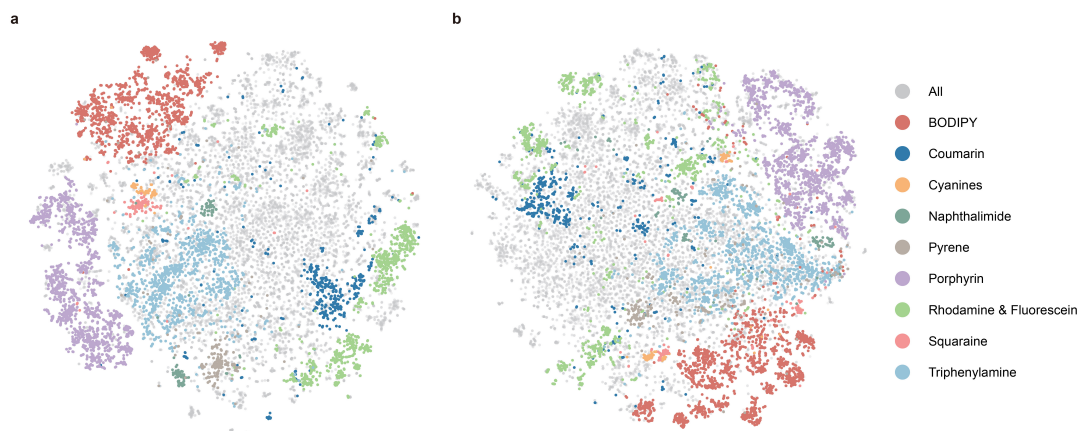

**Fig. S1 | Comparison of latent space visualizations between LUMOS and CDDD.** t-SNE visualizations of the fluorescent chemical space constructed using latent representations from **(a)** LUMOS and **(b)** the CDDD (continuous and data-driven descriptor) model. Colored points correspond to distinct fluorophore scaffolds (e.g., BODIPY, Coumarin) as indicated in the legend, while the gray background represents the full dataset.

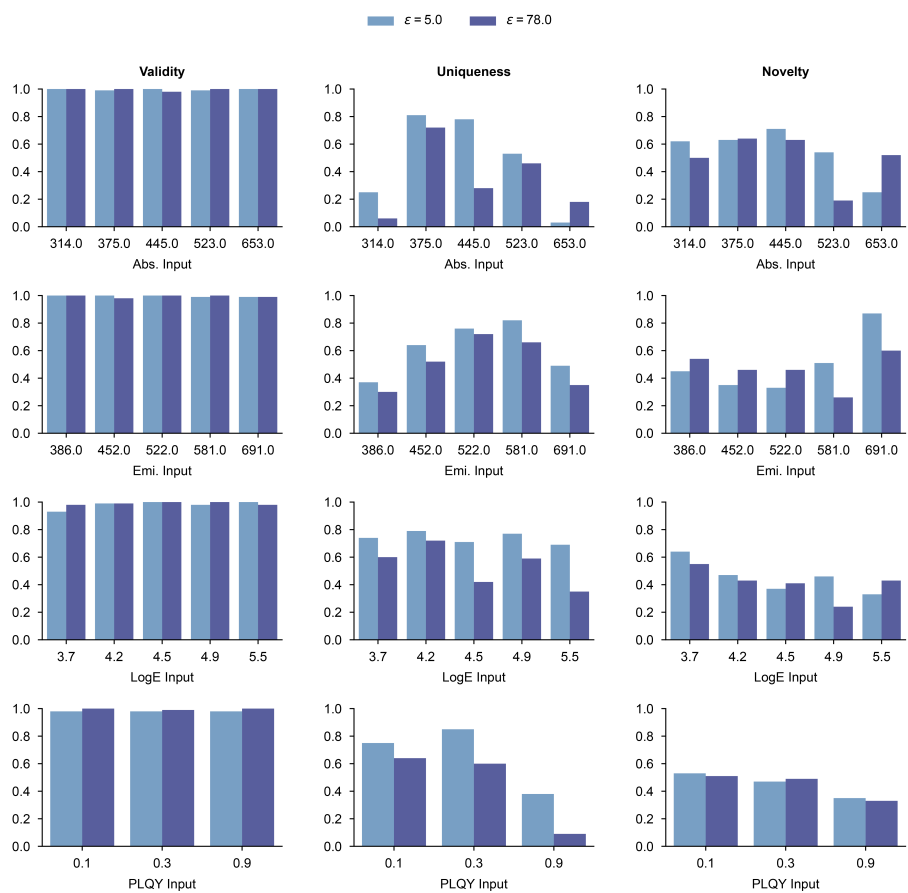

**Fig. S2 | Evaluation metrics of prompt-conditioned *de novo* generation.** The bar plots display validity, uniqueness, and novelty scores across varying input prompts for absorption, emission, $\log \epsilon$ and PLQY. Results are compared between two solvent environments: $\varepsilon = 5.0$ (light blue) and $\varepsilon = 78.0$ (dark blue). The validity is defined as the percentage of generated molecules that pass the sanitization process in RDKit. The uniqueness is defined as the proportion of non-duplicate structures among the valid molecules. The novelty is defined as the proportion of valid, unique molecules that are not in the training dataset.

# Supplementary Tables

**Table S1 | The SMARTS strings for different fluorophores.**

| Fluorophore name | SMARTS |
|---|---|
| BODIPY | [X]-[B]1(-[X])-[*]2~[#6]~[#6]~[#6]~[#6]~2~[#6]~[#6]2~[*]-1~[#6]~[#6]~[#6]~2 |
| Coumarin | [!#6]=[#6]1~[#6]~[#6]~[#6]2:[#6]:[#6]:[#6]:[#6]:[#6]:2~[!#6]~1 |
| Cyanines | [*]-[#7+]1=[#6](-[#6]=[#6]2-[#7](-[*])~[#6]~[#6]~[#6]~2)-[#6]~[#6]~[#6]~1 |
| | [*]-[#7+]1=[#6](-[#6]=[#6]-[#6]=[#6]2-[#7](-[*])~[#6]~[#6]~[#6]~2)-[#6]~[#6]~[#6]~1 |
| | [*]-[#7+]1=[#6](-[#6]=[#6]-[#6]=[#6]-[#6]=[#6]2-[#7](-[*])~[#6]~[#6]~[#6]~2)~[#6]~[#6]~[#6]~1 |
| | [*]-[#7+]1=[#6](-[#6]=[#6]-[#6]=[#6]-[#6]=[#6]-[#6]=[#6]2-[#7](-[*])~[#6]~[#6]~[#6]~2)~[#6]~[#6]~[#6]~1 |
| | [*]-[#7+]1=[#6](-[#6]=[#6]-[#6]=[#6]-[#6]=[#6]-[#6]=[#6]-[#6]=[#6]2-[#7](-[*])~[#6]~[#6]~[#6]~2)~[#6]~[#6]~[#6]~1 |
| | [*]-[#7+]1=[#6](-[#6]=[#6]-[#6]=[#6]-[#6]=[#6]-[#6]=[#6]-[#6]=[#6]-[#6]=[#6]2-[#7](-[*])~[#6]~[#6]~[#6]~2)~[#6]~[#6]~[#6]~1 |
| | [*]-[#7+]1=[#6](-[#6]=[#6]-[#6]=[#6]-[#6]=[#6]-[#6]=[#6]-[#6]=[#6]-[#6]-[#6]=[#6]2-[#7](-[*])~[#6]~[#6]~[#6]~2)~[#6]~[#6]~[#6]~1 |
| | [*]-[#7+]1=[#6](-[#6]=[#6]-[#6]=[#6]-[#6]=[#6]-[#6]=[#6]-[#6]=[#6]-[#6]-[#6]=[#6]-[#6]=[#6]2-[#7](-[*])~[#6]~[#6]~[#6]~2)~[#6]~[#6]~[#6]~1 |
| Naphthalimide | [#8]=[#6]1-[#7,#8]-[#6](=[#8])-[#6]2:[#6]:[#6]:[#6]:[#6]3:[#6]:[#6]:[#6]:[#6]-1:[#6]:2:3 |
| Pyrene | [#6]12:[#6]:[#6]:[#6]3:[#6]:[#6]:[#6]:[#6]4:[#6]:3:[#6]:1:[#6](:[#6]:[#6]:[#6]:2):[#6]:[#6]:4 |
| Porphyrin | [#6]12:[#6]:[#6]3:[#6]:[#6]:[#6](:[#7H]:3):[#6]:[#6]3:[#7]:[#6](-[#6]=[#6]-3):[#6]:[#6]3:[#7H]:[#6](:[#6]:[#6](:[#7]:1)-[#6]=[#6]-2):[#6]:[#6]:3 |
| Rhodamine & Fluorescein | [#6]1:[#6]:[#6]:[#6]2:[#6](-[#8]-[#6]3:[#6]:[#6]:[#6]:[#6]:[#6]:3-[#6]-2):[#6]:1 |
| | [#6]1:[#6]:[#6]2:[#8]:[#6]3:[#6]:[#6]:[#6]:[#6]:[#6]:3:[#6]:[#6]-2:[#6]:[#6]:1 |
| | [#6]1:[#6]:[#6]2:[#8]:[#6]3:[#6]:[#6]:[#6]:[#6]:[#6]:3:[#6]:[#6]:2:[#6]:[#6]:1 |
| Squaraine | [#8]~[#6]1~[#6]~[#6](~[#8])~[#6]~1 |
| | [#8]~[#6]1~[#6](~[#8])~[#6]~[#6]~1 |
| Triphenylamine | [#6]1(-[#7](~[#6]2~[#6]~[#6]~[#6]~[#6]~[#6]~2)-[#6]2:[#6]:[#6]:[#6]:[#6]:[#6]:2):[#6]:[#6]:[#6]:[#6]:[#6]:1 |